\documentclass[lettersize,journal]{IEEEtran}
\usepackage{amsmath,amsfonts,amssymb}
\usepackage{algorithmic}
\usepackage{algorithm}
\usepackage{array}
\usepackage[caption=false,font=normalsize,labelfont=sf,textfont=sf]{subfig}
\usepackage{textcomp}
\usepackage{stfloats}
\usepackage{url}
\usepackage{verbatim}
\usepackage{graphicx}
\usepackage{cite}
\usepackage{hhline}
\usepackage{multirow}
\usepackage{stfloats}
\usepackage{xcolor}
\usepackage{colortbl}
\usepackage{pifont}

\begin{document}

\title{Mamba-based Spatio-Frequency Motion Perception for Video Camouflaged Object Detection}

\author{Xin Li,~~Keren Fu$^{*}$,~~and Qijun Zhao
\thanks{Xin Li, Keren Fu and Qijun Zhao are with the College of Computer Science, Sichuan University, Chengdu 610065, China (e-mail:lixin6092@qq.com; fkrsuper@scu.edu.cn; qjzhao@scu.edu.cn). $^{*}$Corresponding author.}}

\maketitle

\begin{abstract}
Existing video camouflaged object detection (VCOD) methods primarily rely on spatial appearances for motion perception. However, the high foreground-background similarity in VCOD limits the discriminability of such features (e.g. color and texture). Recent studies demonstrate that frequency features can not only compensate for appearance limitations, but also perceive motion through dynamic variations in spectral energy. Meanwhile, the emerging state space model called Mamba enables efficient motion perception in frame sequences with its linear-time long-sequence modeling capability. Motivated by this, we propose Vcamba, a visual camouflage Mamba based on spatio-frequency motion perception that integrates frequency and spatial features for efficient and accurate VCOD. Specifically, by analyzing the spatial representations of frequency components, we reveal a structural evolution pattern that emerges from the ordered superposition of components. Based on this observation, we propose a unique frequency-domain sequential scanning (FSS) strategy to unfold the spectrum. Utilizing FSS, the adaptive frequency enhancement (AFE) module employs Mamba to model the causal dependencies within sequences, enabling effective frequency learning. Furthermore, we propose a space-based long-range motion perception (SLMP) module and a frequency-based long-range motion perception (FLMP) module to model spatio-temporal and frequency-temporal sequences. Finally, the space and frequency motion fusion module (SFMF) integrates dual-domain features into unified motion representation. Experiments show that Vcamba outperforms state-of-the-art methods across 6 evaluation metrics on 2 datasets with lower computation cost, confirming its superiority. Code is available at: https://github.com/BoydeLi/Vcamba.

\end{abstract}

\begin{IEEEkeywords}
Video camouflaged object detection, frequency learning, motion perception, visual Mamba.
\end{IEEEkeywords}

\section{Introduction}
Camouflage is a natural strategy that organisms use to blend into their surroundings to evade predators \cite{stevens2009animal}. Camouflaged object detection (COD) aims to identify, locate, and precisely segment objects that are highly similar to their background and thus hard to detect by humans or machines. COD has various applications, including polyp segmentation \cite{fan2020pranet}, lung infection segmentation \cite{fan2020inf}, agricultural pest detection \cite{fan2021concealed}, industrial defect detection \cite{zhang2022fdsnet}, and underwater species recognition \cite{chen2022robust}. Unlike traditional visual segmentation tasks such as salient object detection (SOD) \cite{jiang2024transformer, mou2024salient, yin2024key}, COD faces the unique challenge of high foreground-background similarity, as objects share similar textures, colors, and patterns with their surroundings. Existing COD methods \cite{zhang2016bayesian, galun2003texture, tankus1998detection, guan2024sdrnet, chen2022boundary, li2025camouflaged} rely on subtle appearance discrepancies between foreground and background, and typically use multi-stream inputs \cite{wu2023source} and multi-task supervision \cite{zhai2021mutual} to enhance segmentation.

However, it is challenging for both humans and machines to accurately localize camouflaged objects based solely on appearances in static scenes. In nature, organisms identify prey or predators by perceiving dynamic changes in the environment, as motion reveals inconsistencies with the static background. This phenomenon aligns with the two-stream hypothesis of neuroscience \cite{goodale1992separate}, which indicates that motion is crucial for breaking camouflage. Based on this, existing video camouflaged object detection (VCOD) methods \cite{cheng2022implicit, hui2024endow, hui2024implicit, zhang2025explicit} break camouflage by explicitly modeling motion cues in frames like optical flow or implicitly learning them through temporal context.

\begin{figure}[t]
\centerline{\includegraphics[scale=0.60, width=1\linewidth]{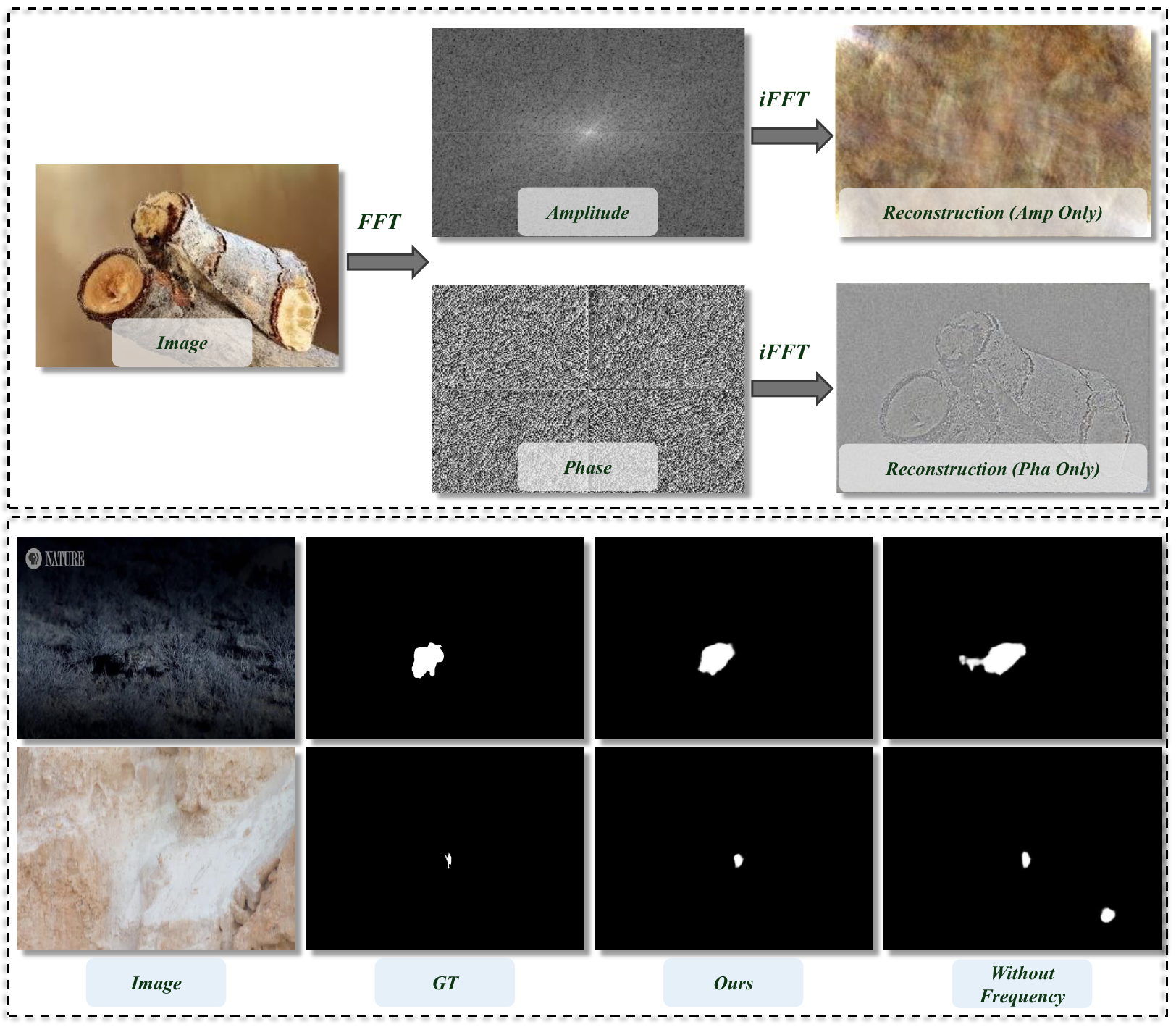}}
\caption{Visualization of frequency phase and amplitude representations in VCOD (top). The input is a camouflaged RGB image from VCOD dataset. Visual Comparison of prediction with and without frequency information (bottom). For the model without frequency, we remove the whole frequency branch and keep only the spatial modeling.}
\label{fig1}
\end{figure}

Although existing methods have made progress, relying solely on spatial appearance features limits performance. The high similarity between foreground and background in texture, color, and pattern results in limited discriminative cues and increased background noise in spatial domain. Recent COD methods \cite{zhang2025frequency,sun2024frequency,zhong2022detecting} have demonstrated the effectiveness of frequency features for enhancing foreground-background distinction. More critically, spatial features are limited by local pixel dependencies and lack global semantic awareness, whereas frequency features inherently provide global representation \cite{chi2020fast, qin2021fcanet}. Additionally, previous studies \cite{cui2009temporal, jia2016saliency} have shown that inter-frame motion can be effectively represented by dynamic variations in frequency-domain energy. Therefore, Integrating frequency features into VCOD supplements discriminative cues and enables global motion perception, compensating for spatial feature limitations. As shown in Fig.~\ref{fig1}, frequency features enable our model to deliver finer detail segmentation and more accurate global localization. 

In addition, most existing COD and VCOD methods \cite{guan2024sdrnet,hui2024implicit,zhang2025frequency} are based on convolutional neural networks (CNNs) and Transformers. However, CNN-based methods struggle to capture global features due to local receptive fields, while Transformer-based methods enable global modeling through self-attention, but facing efficiency issues from quadratic complexity. Since VCOD requires both global perception and low computation cost, the emerging state space model (SSM) called Mamba \cite{gu2111efficiently} enables global modeling in linear-time, making it an ideal fit for VCOD, which has been extended from natural language processing (NLP) to vision tasks for efficient long-range modeling of images after converting them into 1D representations \cite{vim,liu2024vmamba}. It achieves strong performances in downstream visual tasks \cite{liu2024vmamba,gong2025nnmamba,qiao2024vl}.

However, the application of Vision Mamba in VCOD and its synergy with frequency remain underexplored. Directly employing it for encoding fails to fully realize its potential. The core of VCOD lies in breaking camouflage via motion cues. In essence, motion corresponds to object variations along the temporal sequence, which can be manifested both in spatial features and frequency-domain energy. Thus, inter-frame motion perception fundamentally involves modeling long-range dependencies across frame sequences in both spatial and frequency domains, as the interpretation of the current frame relies on accumulated cues from previous frames. Mamba naturally fits such causal dependency modeling.

Meanwhile, directly leveraging frequency features introduces additional noise. Existing methods \cite{zhang2025frequency, sun2024frequency, cong2023frequency} have begun exploring frequency modeling in COD, typically using convolution or self-attention to reweight or correlate spectral components, thereby adaptively focusing on target-relevant components. However, these approaches mainly focus on static correlations between components, without explicitly considering the long-range dependencies among them. In fact, by analyzing the spatial representations of spectral components, we discover that they exhibit a structural evolution pattern during ordered superposition, which aligns closely with the progressive process by which human vision and neural networks uncover camouflaged objects. In contrast, SSMs can continuously pass historical information through hidden states, enabling current output to build upon structures formed by preceding components, which corresponds to the superposition process of frequency components. Therefore, through Mamba, we can perceive long-range dependencies in component sequences and uncover latent structural evolution patterns, achieving effective frequency learning.

Based on all these insights, we propose Vcamba, a Mamba-based spatio-frequency motion perception network. Specifically, We employ VMamba \cite{liu2024vmamba} as the backbone, given its widespread adoption in downstream tasks \cite{he2025samba,chen2025frequency,zhang2025mser}. We propose an adaptive frequency enhancement (AFE) module that introduces a novel frequency-domain sequential scanning (FSS) strategy. This strategy unfolds frequency components in both low-to-high and high-to-low orders. By propagating components sequentially, the SSM enables their ordered superposition at the spatial representation level, thereby capturing latent structural evolution patterns and achieving adaptive frequency enhancement. We further introduce a dual-branch framework composed of a space-based long-range motion perception (SLMP) module and a frequency-based long-range motion perception (FLMP) module. The two branches adopt spatio-temporal and frequency-temporal scanning strategies, respectively, organizing frames into spatio-temporal and frequency-temporal sequences. By modeling spatial structural variations and spectral energy changes of targets over time, they extract complementary dual-domain motion cues. Additionally, we analyze the spatial representations of amplitude and phase in camouflaged scenes. As shown in Fig.~\ref{fig1}, visualizations indicate that amplitude encodes appearance details, while phase emphasizes structure and contours. According to the Fourier shift theorem, spatial displacement induces linear changes in phase while leaving amplitude unchanged \cite{fleet1990computation}. Therefore, FLMP leverages phase variations to perceive structural motion, effectively suppressing appearance interference. Finally, we propose a spatial and frequency motion fusion (SFMF) module, which aggregates dual-domain motion cues via point-to-point and sequence-to-sequence concatenation for unified motion representation. The main contributions are as follows:
\begin{itemize}
\item{We propose a novel visual camouflage Mamba based on spatio-frequency motion perception to address the limited discriminability of spatial appearance features, and achieve efficient and accurate VCOD.}
\item{We propose a frequency-domain sequential scanning strategy that unfolds FFT-derived frequency components in low-to-high and high-to-low spiral order to achieve adaptive frequency component enhancement.}
\item{We propose spatio-temporal and frequency-temporal scanning strategies to perceive dual-domain motion in spatial and frequency phase domains, and introduce a sequence-to-sequence and point-to-point dual-domain sequence concatenation strategy for motion fusion.}
\item{Our Vcamba achieves state-of-the-art performance across six evaluation metrics on two benchmark datasets, and has lower computation cost compared to other VCOD methods, demonstrating the effectiveness and efficiency of our method.}
\end{itemize}

\section{RELATED WORKS}
\subsection{Video Camouflaged Object Detection}
Existing COD methods \cite{fan2020camouflaged,jia2022segment,zhong2022detecting,zhai2021mutual} use CNNs and transformers to extract static visual features like color and texture to break camouflage. Fan \textit{et al.} \cite{fan2020camouflaged} first introduced COD task and established a two-stage search-and-identify network as a baseline. However, static visual features are easily disrupted by camouflage, making them less discriminative. In contrast, dynamic motion cues across video frames are considered crucial for breaking camouflage. Traditional video methods exploit motion cues \cite{conte2009algorithm} or optical flow \cite{hou2011detection} to perceive camouflaged objects. Cheng \textit{et al.} \cite{cheng2022implicit} argued that the ambiguous and indistinct boundaries of camouflaged objects hinder the accuracy of explicit motion estimation methods such as optical flow. To address this, they proposed a Transformer-based network called SLTNet that implicitly captures motion through long-range modeling. Hui \textit{et al.} \cite{hui2024implicit} integrated implicit and explicit motion learning into a unified framework, proposing IMEX, which leverages global frame alignment and explicit motion supervision. Then, they further constructed self-prompts to harness the powerful Segment Anything Model \cite{kirillov2023segment} for precise segmentation in \cite{hui2024endow}. Most of the above methods explore motion cues in the spatial domain. However, discriminative features in the spatial domain are limited due to the foreground-background similarity. We further explore frequency feature and integrate it with spatial motion cues to achieve complementary enhancement.

\subsection{Visual Mamba}
Mamba \cite{gu2023mamba} is a general sequence modeling architecture based on the selective State Space Model (SSM). Its SSM effectively captures long-range dependencies, demonstrating linear time complexity relative to sequence length. This addresses the limitations of Transformers in long-range dependency modeling performance and computational efficiency. Mamba was initially proposed in NLP for language sequences modeling \cite{gu2111efficiently}. Then, the first visual Mamba backbone, Vision Mamba \cite{vim}, was introduced. It utilizes a bidirectional scanning method to serialize images into 1D sequences, with the SSM capturing long-range dependencies in the image sequence. Based on this, Liu \textit{et al.} \cite{liu2024vmamba} proposed the more powerful Vmamba, which performs four-directional scanning in both horizontal and vertical directions. Furthermore, to better adapt Visual Mamba to downstream salient object detection, He \textit{et al.} \cite{he2025samba} proposed a saliency Mamba called Samba, which employs a spatial neighboring scanning strategy to preserve the spatial continuity of salient objects. In addition, Mamba has also been widely applied to image classification \cite{gong2025nnmamba}, semantic segmentation \cite{liu2024vmamba}, and multi-modal learning \cite{qiao2024vl}. However, the application of Mamba in VCOD remains under-explored. Since motion cues are crucial for VCOD and inherently arise from long-range temporal dependencies in frame sequences, we employ Mamba to model both spatio-temporal and frequency-temporal sequences. Furthermore, we also introduce a novel frequency-domain scanning strategy that enables Mamba to perform adaptive frequency learning.

\subsection{Frequency Learning}
Frequency learning is widely used in signal analysis to reveal the time-frequency characteristics and energy distribution patterns of signals. The rich global features contained in the frequency domain can also offer a more comprehensive understanding of image patterns. By applying frequency-domain convolution and other attention mechanisms, it enables frequency learning that filters out irrelevant components and focuses on target-specific component range. The feasibility of frequency learning in COD has been demonstrated. Zhong \textit{et al.} \cite{zhong2022detecting} introduced offline frequency cues and proposed a frequency perceptual loss to learn them. Cong \textit{et al.} \cite{cong2023frequency} used octave convolution to separate high and low frequency features, learning them to obtain coarse localization information. He \textit{et al.} \cite{he2023camouflaged} applied spatial and channel attention to highlight notable frequency parts. Sun \textit{et al.}\cite{sun2024frequency} further leveraged self-attention for global frequency modeling. Liu \textit{et al.} \cite{liu2024edge} proposed to reduce noise in frequency domain by contrastive learning. Zhang \textit{et al.} \cite{zhang2025frequency} employed global average pooling and convolution to generate weights for adaptively learning frequency amplitudes. Previous methods rely on convolution or self-attention for frequency learning, while overlooking the long-range dependencies within frequency component sequences. By analyzing the spatial representations of frequency components, we uncover a latent structural evolution pattern emerging from their ordered superposition, which aligns with the process of object discovery in COD. It can be effectively captured by Mamba’s causal modeling to facilitate more powerful frequency learning.

\begin{figure}[t]
\centerline{\includegraphics[scale=1, width=1\linewidth]{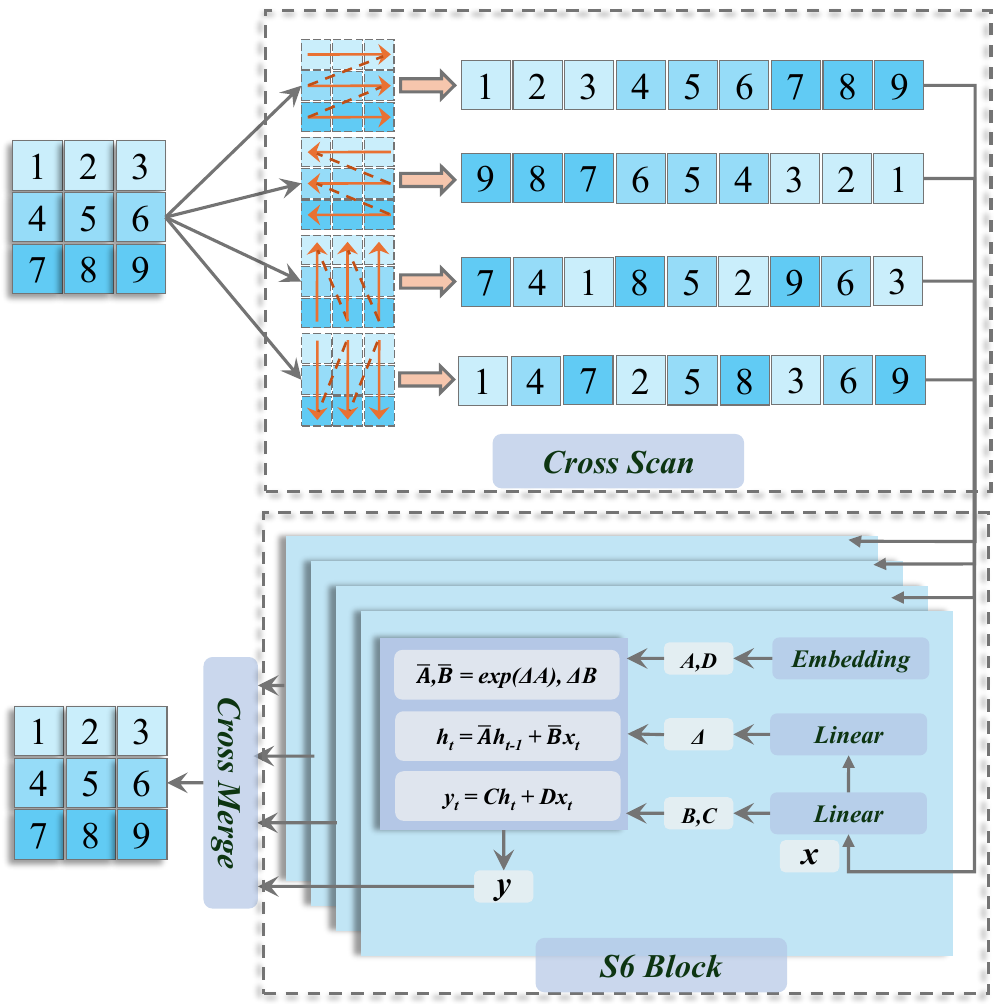}}
\caption{Diagram of the selective scan (SS2D) module, including the details of cross scan and S6 Block.}
\label{fig2}
\end{figure}

\section{METHODOLOGY}
\subsection{Preliminaries} \label{section III A}
\textbf{State Space Model.} SSMs \cite{gu2111efficiently} originate from the Kalman filter \cite{kalman1960new} and can be regarded as linear time-invariant (LTI) systems. They map a one-dimensional sequence or function $x\left( t \right) \in \mathbb{R}$ to $y\left( t \right) \in \mathbb{R}$ through a hidden state $h\left( t \right) \in \mathbb{R}^N$, and can be formulated as the following linear ordinary differential equations (ODEs):
\begin{equation}
y\left( t \right) =Ch\left( t \right) +Dx\left( t \right) ,h'\left( t \right) =Ah\left( t \right) +Bx\left( t \right) ,\label{eq1}
\end{equation}
where $A\in \mathbb{R}^{N\times N}$ is the state transition matrix, $B\in \mathbb{R}^{N\times 1}$ and $C\in \mathbb{R}^{1\times N}$ are projection matrices, and $D\in \mathbb{R}^{1}$ is a skip connection. $h'\left( t \right)$ denotes the time derivative of $h\left( t \right)$. To adapt to discrete sequence data and integrate into deep models, SSMs are discretized via zero-order hold (ZOH) \cite{gu2111efficiently} to convert the continuous system into a discrete form, as shown below:
\begin{equation}
\bar{A}=\exp \left( \varDelta A \right) ,\bar{B}=\left( \varDelta A \right) ^{-1}\left( \exp \left( A \right) -I \right) \cdot \varDelta B,\label{eq2}
\end{equation}
where $\varDelta$ is the timescale parameter. Then the discretized SSMs can perform sequence modeling through recursive computation, mapping the input sequence $x\left( t \right)$ to the output sequence $y\left( t \right)$ via the following equations:
\begin{equation}
h_t=\bar{A}h_{t-1}+\bar{B}x_t,y_t=Ch_t+Dx_t,\label{eq3}
\end{equation}

The structured state space model (S4) achieves linear time complexity, but its time-invariant nature where parameters $A$, $B$, $C$, and $D$ are static limits its ability to dynamically capture sequence context. To address this, Mamba \cite{gu2023mamba}, based on the selective state space model (S6), was proposed. It introduces a dynamic, input-dependent parameterization mechanism in which $B$, $C$, and $\varDelta$ are directly generated from the input sequence, enabling adaptive selection of key information. Additionally, Mamba incorporates hardware-aware parallel scan optimization for more efficient sequence modeling.

\textbf{Cross Scan in VMamba.} Traditional Mamba focuses on 1D sequence modeling and struggles to handle 2D spatial information in images. To address this, Vision Mamba \cite{vim} proposes a bidirectional scanning mechanism that unfolds the image into forward and backward sequences along the horizontal direction. As shown in Fig.~\ref{fig2}, VMamba \cite{liu2024vmamba} introduces 2D selective scan using a cross-scan strategy along four directional paths, corresponding to the vertical, horizontal and their reverse directions. Each of the four sequences is processed through a S6 module for long-range dependency modeling. Subsequently, the sequences are reshaped back into 2D space via cross merge, achieving more comprehensive and effective spatial modeling.

\textbf{Phase-based Motion Perception.} According to the Fourier shift theorem \cite{oppenheim2005importance}, a spatial translation introduces a linear phase shift in the frequency domain while leaving the amplitude unchanged, and phase is more sensitive and robust than amplitude to contrast variations and affine deformations \cite{fleet1990computation}. Based on this principle, Hommos et al. \cite{hommos2018using} showed that phase variations directly correspond to object motion, making phase a viable representation for motion modeling. Phase information has also been widely exploited in motion representation and video processing \cite{wadhwa2013phase,zhou2010phase,pintea2016making}.

Specifically, for an image $f(x, y)$, its 2D Fourier transform is denoted as $F(u,v)$:
\begin{equation}
F(u, v) = \iint f(x, y)\, e^{-j 2\pi (u x + v y)} \, dx \, dy,
\label{eq5}
\end{equation}
after spatial translation by $( \Delta x, \Delta y)$, 
\begin{equation}
g(x, y) = f(x - \Delta x,\, y - \Delta y),
\label{eq6}
\end{equation}
the transformed spectrum becomes $G(u,v)$:
\begin{equation}
G(u, v) = F(u, v)\, e^{-j 2\pi (u \Delta x + v \Delta y)},
\label{eq7}
\end{equation}
specifically, the amplitude is obtained by applying the modulus operation $|\cdot|$ to $G(u,v)$, while the phase is computed by taking the argument (i.e., the angle) of the complex spectrum using the arctangent function $\angle(\cdot)$:
\begin{equation}
  \begin{aligned}
    & |G(u, v)| = \left| F(u, v)\, e^{-j 2\pi (u \Delta x + v \Delta y)} \right|
           = |F(u, v)|,\\
    & \angle G(u, v) = \angle F(u, v) - 2\pi (u \Delta x + v \Delta y),
   \end{aligned}
  \label{eq8}
\end{equation}
Where $|G(u, v)|$ denotes the amplitude of $G(u,v)$, the translation introduces a complex exponential term whose modulus equals one, leaving the amplitude spectrum unchanged. $\angle G(u, v)$ denotes the phase of $G(u,v)$, which undergoes a linear shift with respect to the phase of $F(u,v)$.

\begin{figure*}[t]
\centerline{\includegraphics[width=0.8\linewidth]{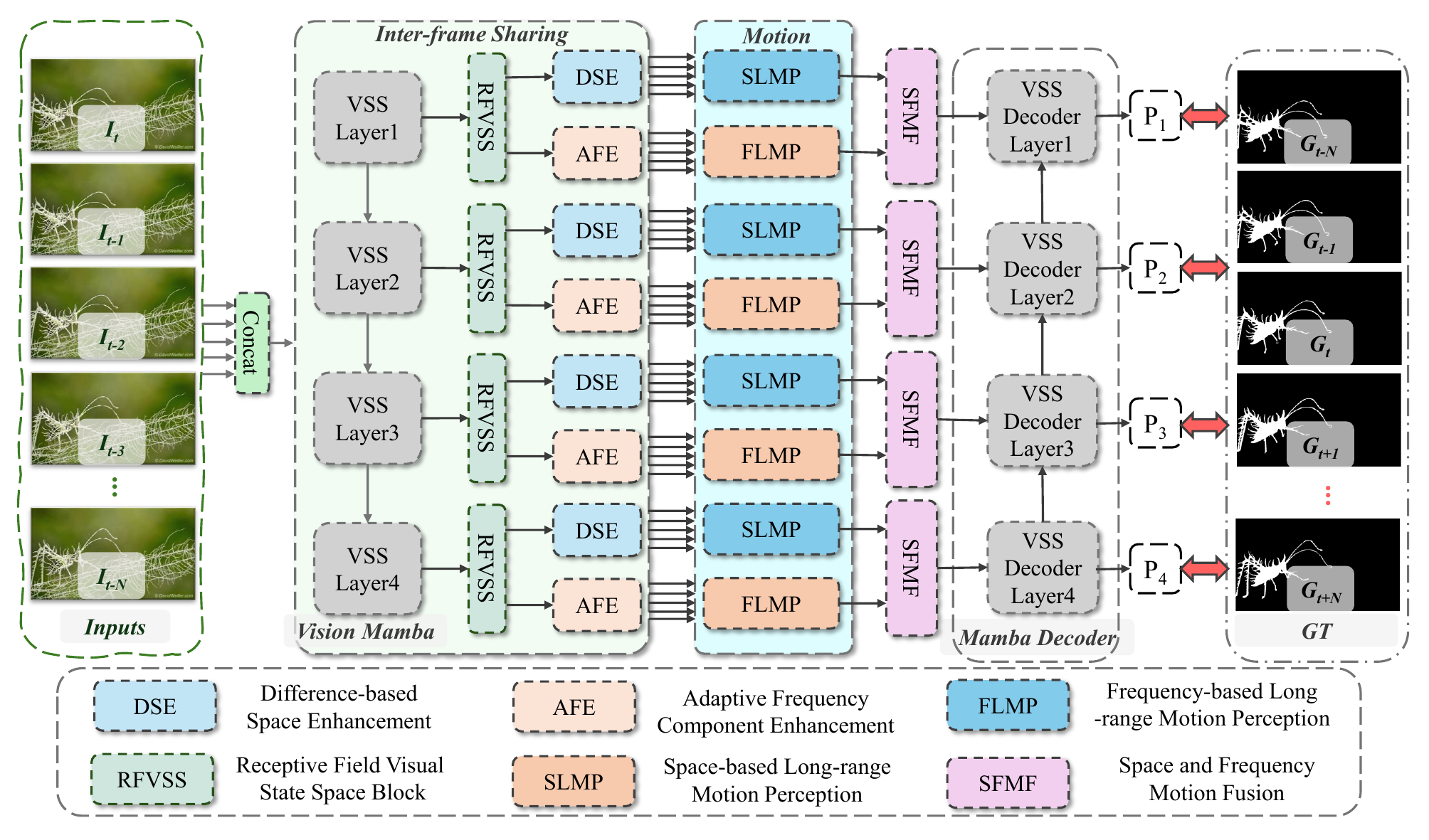}}
\caption{Overview of the proposed Vcamba. $I_t$ denotes to the input, which concatenates the windowed sequence containing N frames, $P_i$ represents the hierarchical prediction of our model, and $G_t$ denotes the ground truth corresponding to each input frame. Each VSS Decoder Layer consists of one VSSBlock.}
\label{fig3}
\end{figure*}

\subsection{Overview}
The overall architecture of the Vcamba is shown in Fig. ~\ref{fig3}. It adopts a U-shaped structure. A video $\left\{ I_t \right\} _{t=1}^{N},I_t\in \mathbb{R}^{3\times H\times W}$ with $N$ frames is encoded by ImageNet pre-trained VMamba to extract hierarchical features $ \left\{ F_i \right\} _{i=1}^{4},F_i\in \mathbb{R}^{N\times 3\times H\times W}$. 

As receptive field block (RFB) \cite{liu2018receptive} is widely used in SOD and COD \cite{fan2020camouflaged,li2025camouflaged,fan2021concealed}, we propose a receptive field VssBlock (RFVSS) to compensate for multi-scale local context in global feature representations from vision Mamba. As shown in the right part of Fig. ~\ref{fig3}, RFVSS employs a receptive field feed-forward network (RF-FFN) that first expands channel dimensions ($C$ to $ 4\times C$) through a linear layer with GELU activation, then extracts multi-scale features via parallel convolutions with varying kernel sizes. To enable efficient multi-scale extraction from high-dimensional features enriched with latent cues,  we replace standard convolutions with lightweight depth-wise separable ones. With RF-FFN, multi-scale context awareness can be directly incorporated into VSSBlock, addressing the deficiency of SSM in capturing local multi-scale details.

These features are then fed into a dual-branch structure for frequency and spatial motion perception. In each branch, domain-specific features are enhanced for long-range motion modeling. For the frequency branch, we propose an adaptive frequency component enhancement (AFE) module employing a novel frequency-domain sequential scanning strategy to adaptively enhance components.  We further propose a frequency-based long-range motion perception (FLMP) module that performs a frequency-temporal scanning strategy in the phase domain to eliminate interference from appearances. 

For the spatial branch, inspired by \cite{pang2024zoomnext}, we propose a difference-based space enhancement (DSE) module that extracts inter-frame differences $ \left\{ F_{i}^{d} \right\} _{i=1}^{N}$ via temporal shift, followed by intra-frame self-attention:
\begin{equation}
  \begin{aligned}
     & F_{i}^{d}=F_{i+1}-F_i,\\
     & F_{i}^{se}=F_{i}^{d}W_Vsoft\max \left( \frac{\left( F_{i}^{d}W_K \right) ^TF_{i}^{d}W_Q}{\sqrt{HW}} \right)
  \end{aligned}
 \label{eq4}
\end{equation}
spatial enhancement features $ \left\{ F_{i}^{se} \right\} _{i=1}^{N}$ are then unfolded with our spatio-temporal scanning strategy to achieve space-based long-range motion perception (SLMP). 

The dual-domain motion features are then passed through the space and frequency motion fusion module (SFMF) module to achieve both global and local integration, alleviating domain and structural discrepancies. The unified motion representations are decoded through a VssBlock-based decoder, with supervision from binary masks of input frames $ \left\{ M_t \right\} _{t=1}^{N},M_t\in \mathbb{R}^{1\times H\times W}$.

\begin{figure*}[t]
\centerline{\includegraphics[scale=1, width=0.8\linewidth]{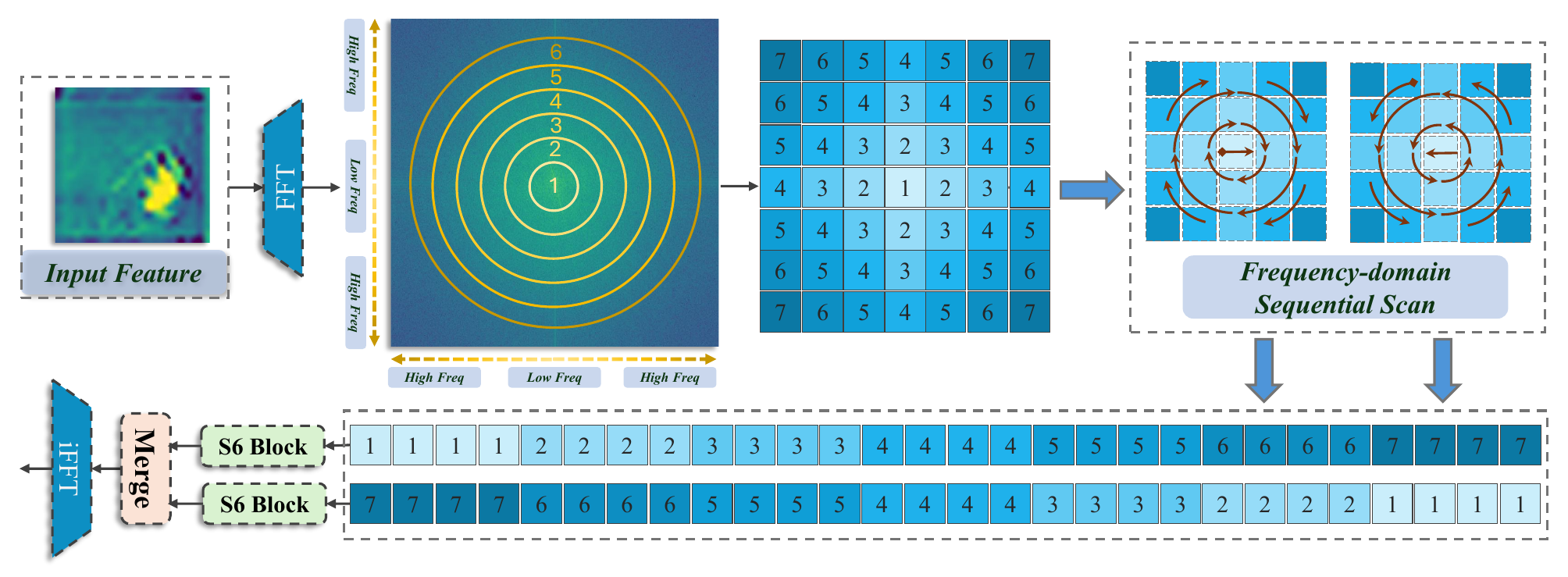}}
\caption{Diagram of the adaptive frequency component enhancement (AFE) module and the frequency-domain sequential scanning (FSS) strategy.}
\label{fig4_1}
\end{figure*}

\subsection{Adaptive Frequency Component Enhancement}
Due to the inherent concealment and ambiguity, spatial modeling of camouflaged objects based on local pixel dependencies often lacks sufficient discriminability and global context. Frequency-domain representations provide a complementary global view, where different spectral components encode spatial representations at distinct granularities. Previous studies \cite{liu2024edge,xie2023frequency,zhong2022detecting} have shown that high-frequency components capture fine-grained spatial details while low-frequency components encode structural information. Since camouflaged object cues and background noise coexist across components, effective COD requires both low-frequency structural modeling and high-frequency detail perception. Existing frequency-based COD methods \cite{zhang2025frequency,sun2024frequency,cong2023frequency} have explored adaptive modulation of components or inter-component correlations. However, these methods mainly focus on component-wise reweighting or correlation modeling, without perceiving long-range dependencies or exploiting the underlying dependency pattern across components.

Since long-range dependencies are inherently order-sensitive, and frequency learning aims to modulate the spatial representations of components, we seek to mine the latent dependency order of spatial representations. Inspired by U-Net \cite{ronneberger2015u}, where object representations are progressively formed via hierarchical feature superposition, the ordered superposition of components naturally aligns with this structural evolution process. Consequently, we propose the frequency-domain sequential scanning (FSS) strategy.

FSS unfolds the spectrum into continuous low-to-high and high-to-low sequences. As shown in Fig.~\ref{fig4_1}, spatial features are transformed into the frequency domain via 2D FFT. The low-to-high sequence starts from the spectral center and spirals outward along increasing radii in a clockwise manner, while the high-to-low sequence is obtained by reversing this order. Since SSMs model \textit{causal} sequences by passing historical information through hidden states, each output depends on preceding inputs in a causal manner. This mechanism closely mirrors the ordered superposition process of frequency components. The two bi-directional sequences are thus processed by parallel S6 modules to perceive long-range dependencies and exploit structural evolutionary patterns, resulting in enhanced frequency representations. The enhanced frequency features are then transformed back to the spatial domain via iFFT for subsequent motion perception.

\begin{figure*}[t]
\centerline{\includegraphics[width=0.8\linewidth]{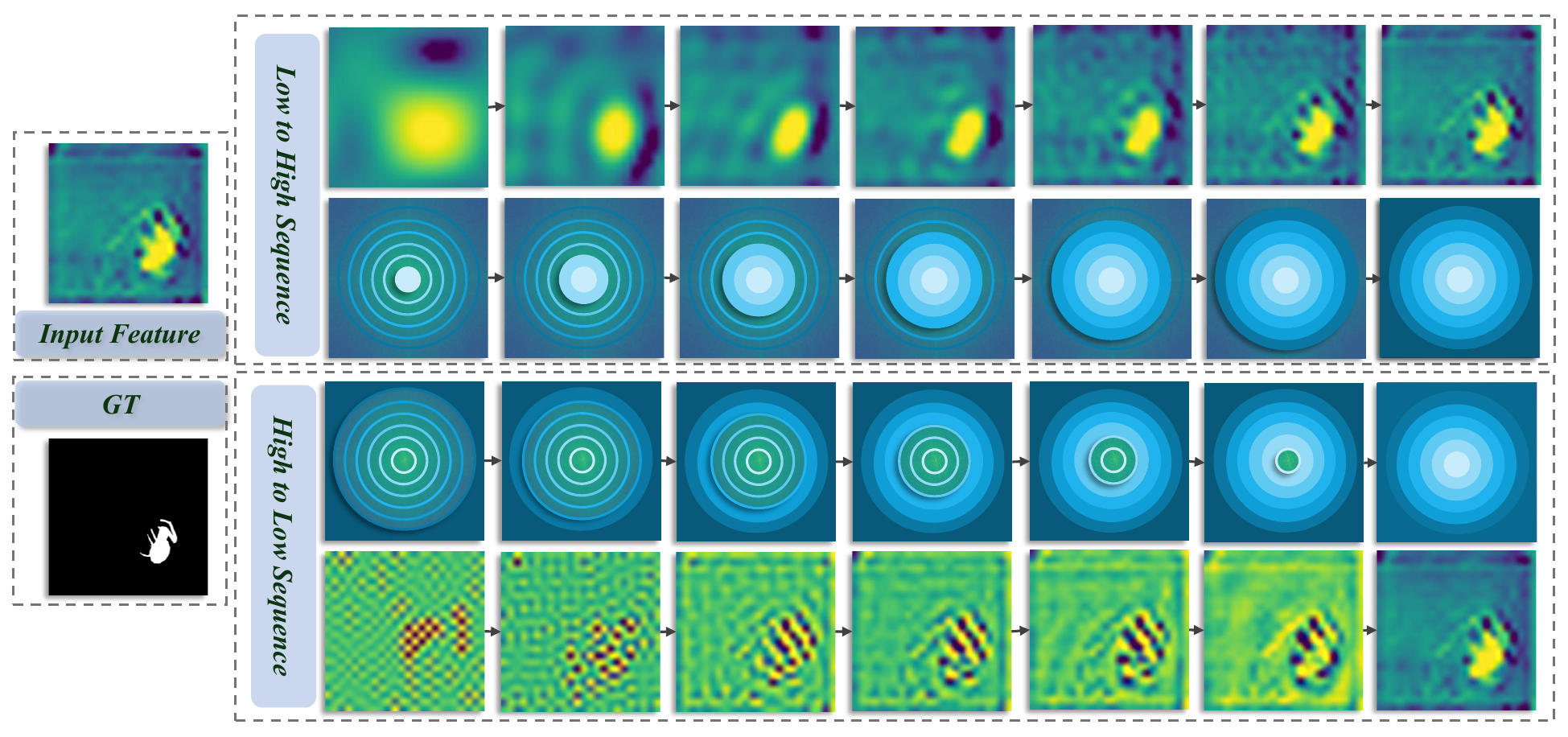}}
\caption{Diagram of the structural evolution pattern emerging from the ordered superposition of frequency components. The low-to-high and high-to-low sequences are constructed by superposing components unfolded through the FSS. We provide a schematic illustration of the superposition process for each sequence, along with their corresponding spatial features.}
\label{fig4_2}
\end{figure*}

In order to illustrate the rationale of FSS, we visualize the ordered superposition of frequency components in Fig.~\ref{fig4_2}. By sequentially superposing frequency components and reconstructing their spatial representations, one can see that the low-to-high sequence simulates the natural visual process from coarse perception to local refinement, which closely matches the coarse-to-fine paradigm of U-Net. Conversely, one can see that the high-to-low sequence demonstrates an evolution from fine details to holistic structures. Both sequences embody a structural evolution pattern with causal relations, and by modeling such long-range dependencies, SSMs enable more effective frequency learning, thereby facilitating accurate object detection.

\begin{figure}[t]
\centerline{\includegraphics[scale=1, width=1\linewidth]{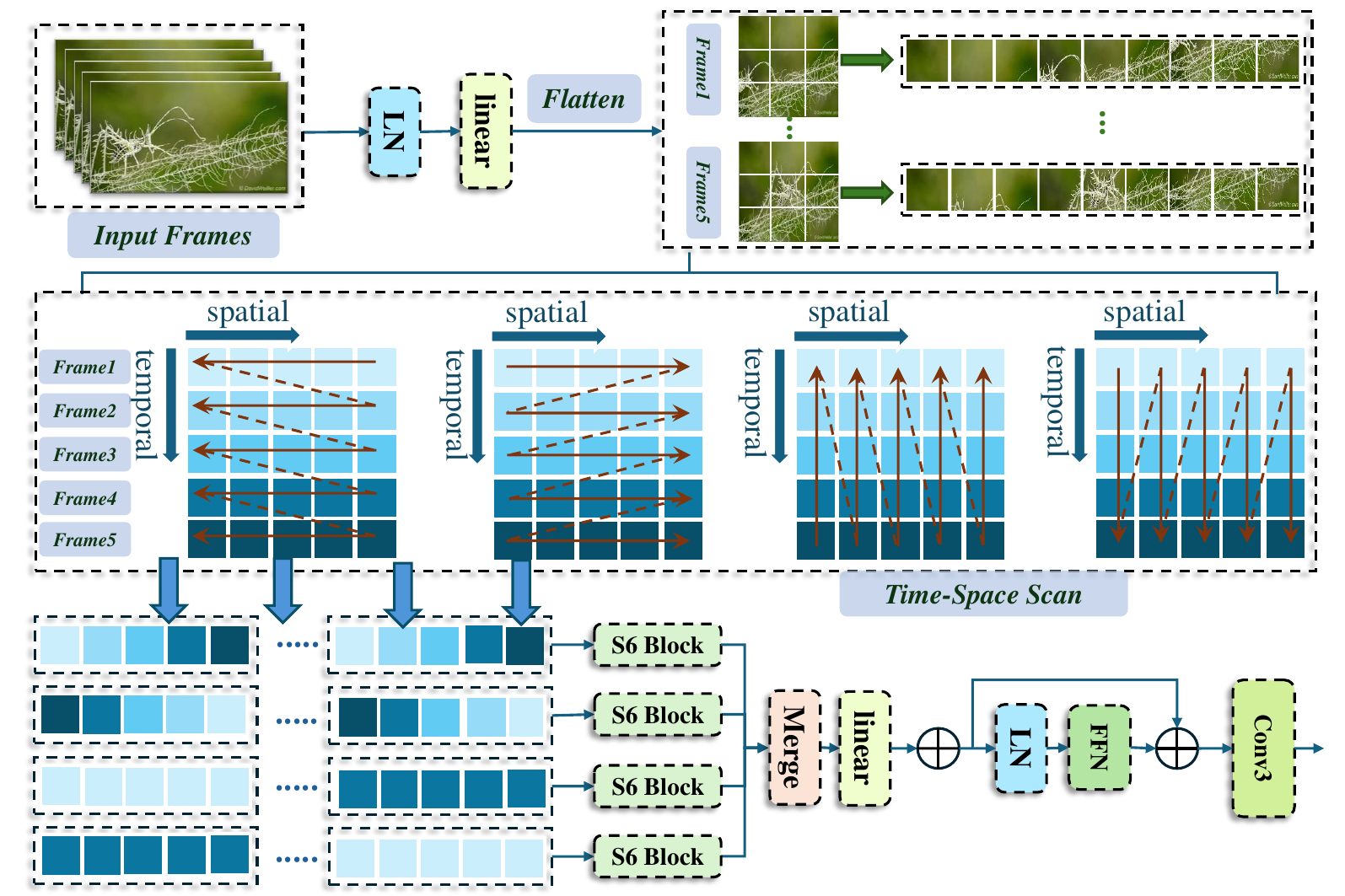}}
\caption{Diagram of the space-based long-range motion perception (SLMP).}
\label{fig5}
\end{figure}

\subsection{Dual Domain Motion Perception}
The core challenge of VCOD lies in how to effectively leverage temporal motion information to break camouflage. We adopt Mamba for motion modeling due to its efficient sequence modeling with linear time complexity and the selective scan of S6 module, which adaptively identifies key regions through dynamic parameterization. In addition, previous studies \cite{cui2009temporal,jia2016saliency,hui2024endow} have shown that inter-frame motion can be represented by dynamic variations in frequency-domain energy. Motivated by this, we propose a novel dual-domain motion perception framework, perceiving motion cues in both frequency and spatial domains.

\textbf{Space-based Long-range Motion Perception.} Since Mamba perceives spatial positions in images by unfolding them into 1D sequences, and the long-range dependencies depend on temporal order of the sequence, as shown in Fig.~\ref{fig5}, we propose a spatio-temporal scanning strategy that serializes video frames along the temporal dimension, enabling long-range dependencies modeling of  spatio-temporal features. 

Specifically, we serialize the frames $\left\{ F_i \right\} _{i=1}^{N},F_i\in \mathbb{R}^{C\times H\times W}$ into 1D sequences $\left\{ S_t \right\} _{t=1}^{N},S_t\in \mathbb{R}^{C\times HW}$  via row-wise flattening, then reconstruct the sequences from different frames along the temporal dimension $N$ into a 2D spatio-temporal map $M_s \in \mathbb{R}^{C \times N \times HW}$. Inspired by cross scan \cite{liu2024vmamba}, we apply four directional scans (horizontal, vertical, and their reverses) to unfold the spatio-temporal map into four sequences. The horizontal one describes global temporal variation across frames, so S6 module can model global dependencies between adjacent frames to perceive comprehensive motion cues. while the vertical one describes the temporal variation of local details at the same spatial location. By modeling inter-frame correlations of these details, the S6 module mines fine-grained motion cues at the local range. Each sequence is modeled by a S6 module and reshaped into 2D feature map via cross merge, followed by FFN and convolution for spatial enhancement. This allows SLMP to leverage Mamba’s sequence modeling for both global and local spatio-temporal motion cues perception$\left\{ F_{i}^{spa} \right\} _{i=1}^{N},F_{i}^{spa}\in \mathbb{R}^{C\times H\times W}$.

\begin{figure}[t]
\centerline{\includegraphics[scale=1, width=1\linewidth]{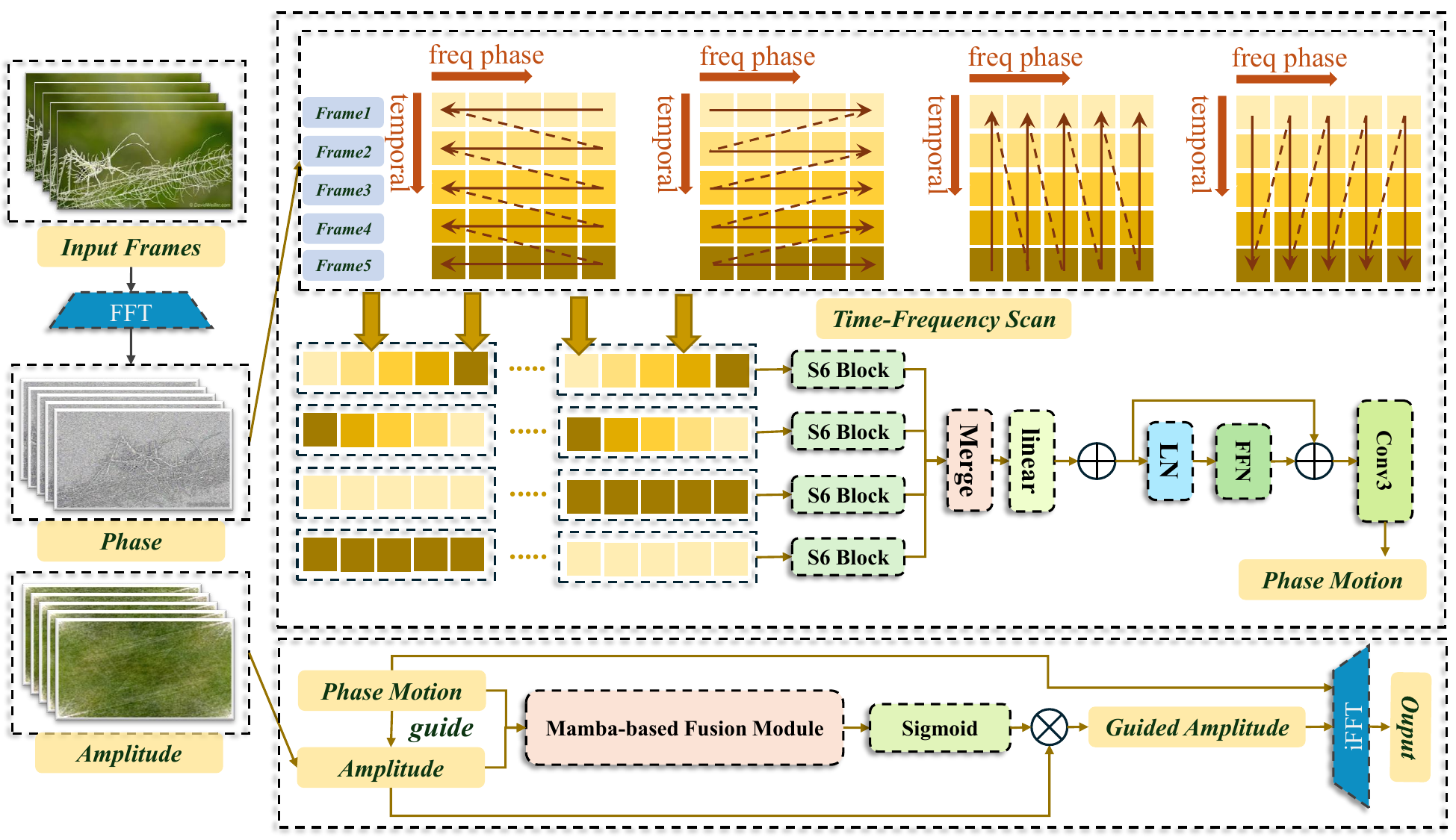}}
\caption{Diagram of the frequency-based long-range motion perception (FLMP).}
\label{fig6}
\end{figure}

\textbf{Frequency-based Long-range Motion Perception.} As shown in Fig.~\ref{fig6}, FLMP follows a similar process to SLMP. In addition, to further explore the potential of frequency in VCOD, as shown in Fig.~\ref{fig1}, we apply FFT to obtain amplitude and phase spectrums of VCOD images, and then use iFFT to reconstruct amplitude-only and phase-only images in spatial domain. As we can see, amplitude represents appearances such as texture and color, while the phase encodes image layout and structural information. Directly leveraging frequency variations to perceive motion may introduce background distraction which is susceptible in VCOD.

The phase can mitigate this by filtering out appearance distractions and enhancing structural perception, making it ideal for motion representation. In addition, as shown in section \ref{section III A} , based on the Fourier shift theorem \cite{oppenheim2005importance}, object motion corresponds to phase variation in the Fourier spectrum while leaving amplitude intact. Compared to amplitude, phase exhibits stronger robustness and sensitivity in perceiving such variations.

Therefore, we only use phase spectrum as carrier for motion perception. Specifically, the spatial features $\left\{ F_i \right\} _{i=1}^{N},F_i\in \mathbb{R}^{C\times H\times W}$ are converted to phase spectrum $\left\{ F_{i}^{_{Pha}} \right\} _{i=1}^{N},F_{i}^{Pha}\in \mathbb{R}^{C\times H\times W}$ and amplitude spectrum $\left\{ F_{i}^{_{Amp}} \right\} _{i=1}^{N},F_{i}^{Amp}\in \mathbb{R}^{C\times H\times W}$ via FFT.  The phase are then flattened and temporally stacked into spatio-frequency map $M_f\in \mathbb{R}^{C\times N\times HW}$. Four-directional sequential modeling is performed on $M_f$ to extract phase motion features $\left\{ F_{i}^{_{Pm}} \right\} _{i=1}^{N},F_{i}^{Pm}\in \mathbb{R}^{C\times H\times W}$. This can be formulated as:

\begin{equation}
  \begin{aligned}
    & F_{i}^{Pha},F_{i}^{Amp}=FFT\left( F_i \right) ,\\
    & M_f=Stack\left( \left\{ Flatten\left( F_{i}^{Pha} \right) \right\} \right) ,\\
    & F_{i}^{Pm}=SSM\left( M_f \right) ,
   \end{aligned}
  \label{eq9}
\end{equation}

While phase motion features $F_{i}^{Pm}$ provide precise structural cues, amplitude $F_{i}^{Amp}$ contain essential image details. A complete frequency representation thus requires combining both. To this end, we employ $F_{i}^{Pm}$ to guide $F_{i}^{Amp}$, filtering out background interference before integration. Specifically, they are fed into a Mamba-based fusion module (MFM) with the same structure as SFMF, allowing amplitude spectrum filtering to be modulated by phase-based motion features at both global and local levels. The phase-guided amplitude maps  $\left\{ F_{i}^{g} \right\} _{i=1}^{N},F_{i}^{g}\in \mathbb{R}^{C\times H\times W}$ are generated by first applying a sigmoid function to the filtered amplitude features, and then multiplying them with the original amplitude spectrum. Finally, $F_{i}^{g}$ and $F_{i}^{Pm}$ are merged via iFFT to produce spatio-frequency motion features $\left\{ F_{i}^{fre} \right\} _{i=1}^{N},F_{i}^{fre}\in \mathbb{R}^{C\times H\times W}$. The entire process can be expressed as:
\begin{equation}
  \begin{aligned}
    & F_{i}^{g}=\sigma \left( MFM\left( F_{i}^{Pm},F_{i}^{Amp} \right) \right) \times F_{i}^{Amp},\\
    & F_{i}^{fre}=iFFT\left( F_{i}^{g},F_{i}^{Pm} \right) ,
   \end{aligned}
  \label{eq10}
\end{equation}
where $\sigma \left( \cdot \right) $ denotes the sigmoid function.  

\begin{figure}[t]
\centerline{\includegraphics[scale=1, width=1\linewidth]{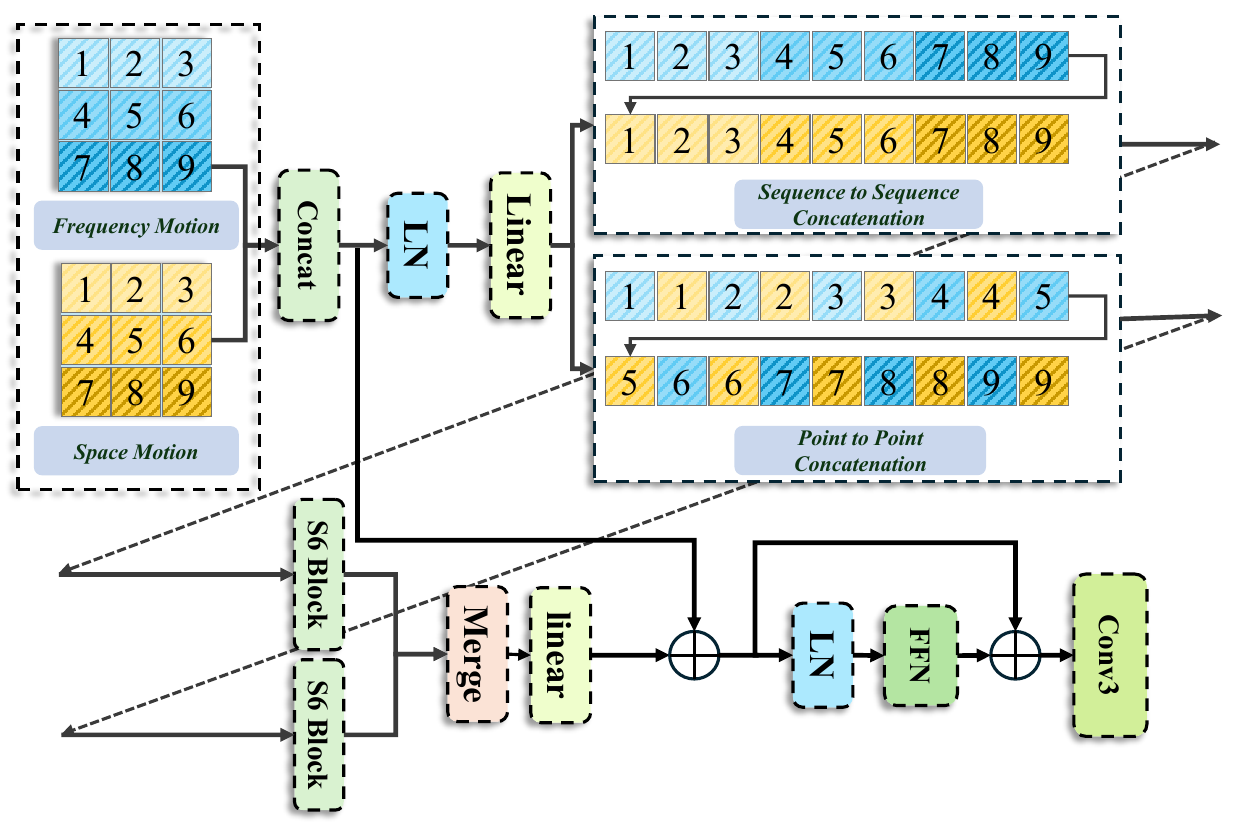}}
\caption{Diagram of the space and frequency motion fusion (SFMF) module.}
\label{fig7}
\end{figure}

\subsection{Space and Frequency Motion Fusion} \label{section III F}
Spatial and frequency domains capture fundamentally different information, which makes dual-domain motion features intrinsically exhibit domain discrepancies, manifesting as distinct representations in global structures and local details. Meanwhile, since the limited discriminability of single-domain features may introduce domain-specific noise, effective fusion of dual-domain motion features can simultaneously enhance complementary regions and mutually suppress such noise. Conventional fusion approaches like local convolutions or global cross-attention often lose critical features while failing to distinguish such noise. To achieve mutual complementarity for more complete and accurate motion information, we propose the space and frequency motion fusion (SFMF) module.

As shown in Fig.~\ref{fig7}, the spatial motion features $\left\{ F_{i}^{spa} \right\} _{i=1}^{N},F_{i}^{spa}\in \mathbb{R}^{C\times H\times W}$ and the frequency motion features $\left\{ F_{i}^{fre} \right\} _{i=1}^{N},F_{i}^{fre}\in \mathbb{R}^{C\times H\times W}$ are unfolded via row-wise flattening. Then, they are concatenated in two manners: sequence-to-sequence and point-to-point. 

The sequence-to-sequence concatenation directly connects dual-domain sequences end-to-end, forming a globally concatenated sequence shaped as $\mathbb{R}^{C\times 2HW}$. The point-to-point concatenation connects patches at identical spatial positions from both domains, forming a locally concatenated sequence shaped as $\mathbb{R}^{C\times 2HW}$. Due to the recursive and decaying nature of state updates in SSMs, each patch in the sequence depends on all preceding patches, especially the nearest one. From a macro perspective, global concatenation establishes a cross-domain causal sequence, where features in subsequent domain holistically depend on those in preceding domain, achieving global contextual fusion. Conversely, local concatenation leverages strong causal correlations between adjacent patches, enabling mutual perception of cross-domain features at identical spatial location. This establishes pixel-level structural correspondences for fine-grained fusion.

Subsequently, the two sequences are reshaped via cross-merge into a concatenated 2D feature map shaped as $\mathbb{R}^{C\times 2H\times W}$, which is further processed by a linear layer and FFN to encode and exchange spatial features.  Finally, a $3\times 3$ convolution integrates the concatenated features into the final unified motion representations of shape $\mathbb{R}^{C\times H\times W}$.

\subsection{Model Optimization}
Following SLTNet \cite{cheng2022implicit}, we adopt a hybrid loss for end-to-end training, which is widely used in VCOD. The hybrid loss is defined as follows:
\begin{equation}
    L_{hybrid}=L_{ce}^{w}+L_{iou}^{w}+L_e \label{eq7}
\end{equation}
where $L_{ce}^{w}$ denotes weighted binary cross-entropy loss \cite{wei2020f3net}, $L_{iou}^{w}$ denotes weighted intersection-over-union \cite{wu2021modality}, and $L_e$ denotes enhanced-alignment loss. As our model adopts multi-frame input and hierarchical prediction, we apply the hybrid loss across all layers and frames. Let $P_{i}^{j}$ denote the prediction for the $j$-th frame at the $i$-th layer, and the total loss is:
\begin{equation}
L=\sum{_{i=1}^{4}}\sum{_{j=1}^{N}\left( L_{hybrid}\left( P_{i}^{j},G^j \right) \right)} \label{eq8}
\end{equation}
where $G^j$ denotes the ground truth of $j$-th frame. Our model sets the number of input frames $N$ to 5.

\begin{table*}[t]
\caption{Quantitative comparison with the 14 state-of-the-art methods for VCOD on two benchmarks. Best video results are highlighted in \textbf{BOLD}. “$\uparrow$ / $\downarrow$” indicates the larger or smaller is better.}
\centering
\resizebox{\linewidth}{!}{
\begin{tabular}{c|c|c|c|c|cccccc|cccccc} 
\hline
\multirow{2}{*}{Method} & \multirow{2}{*}{Year} & \multirow{2}{*}{Input} & \multirow{2}{*}{Backbone} & \multirow{2}{*}{MACs} & \multicolumn{6}{c|}{MoCA-MASK} & \multicolumn{6}{c}{CAD2016}\\ 
\hhline{~~~~~------------}& & & & & {\cellcolor[rgb]{0.89,0.89,0.89}}$S_{\alpha}\uparrow$ & {\cellcolor[rgb]{0.89,0.89,0.89}}$F_{\beta}^{\omega}\uparrow$ & {\cellcolor[rgb]{0.89,0.89,0.89}}$E_{\phi}\uparrow$ & {\cellcolor[rgb]{0.89,0.89,0.89}}$M\downarrow$ & 
{\cellcolor[rgb]{0.89,0.89,0.89}}$mDice\uparrow$ & {\cellcolor[rgb]{0.89,0.89,0.89}}$mIoU\uparrow$ &
{\cellcolor[rgb]{0.89,0.89,0.89}}$S_{\alpha}\uparrow$ & {\cellcolor[rgb]{0.89,0.89,0.89}}$F_{\beta}^{\omega}\uparrow$ & {\cellcolor[rgb]{0.89,0.89,0.89}}$E_{\phi}\uparrow$ & {\cellcolor[rgb]{0.89,0.89,0.89}}$M\downarrow$ & 
{\cellcolor[rgb]{0.89,0.89,0.89}}$mDice\uparrow$ & {\cellcolor[rgb]{0.89,0.89,0.89}}$mIoU\uparrow$\\ 
\hline
\noalign{\vspace{2pt}}
SINet \cite{fan2020camouflaged} & 2020-CVPR & Image & ResNet-50 & 19.42G & 0.574 & 0.185 & 0.655 & 0.030 & 0.221 & 0.156 & 0.601 & 0.204 & 0.589 & 0.089 & 0.289 & 0.209 \\
SINet-v2 \cite{fan2021concealed} & 2021-TPAMI & Image & Res2Net-50 & 12.28G & 0.571 & 0.175 & 0.608 & 0.035 & 0.211 & 0.153 & 0.544 & 0.181 & 0.546 & 0.049 & 0.170 & 0.110 \\
ZoomNet \cite{pang2022zoom} & 2022-CVPR & Image & ResNet-50 & 95.50G & 0.582 & 0.211 & 0.536 & 0.033 & 0.224 & 0.167 & 0.587 & 0.225 & 0.594 & 0.063 & 0.246 & 0.166 \\
BGNet \cite{sun2022bgnet} & 2022-IJCAI & Image & Res2Net-50 & 58.45G & 0.590 & 0.203 & 0.647 & 0.023 & 0.225 & 0.167 & 0.607 & 0.203 & 0.666 & 0.089 & 0.345 & 0.256 \\
FEDERNet \cite{he2023camouflaged} & 2023-CVPR & Image & ResNet-50 & 18.01G & 0.555 & 0.158 & 0.542 & 0.049 & 0.192 & 0.132 & 0.607 & 0.246 & 0.725 & 0.061 & 0.361 & 0.257 \\
FSPNet \cite{huang2023feature} & 2023-CVPR & Image & ViT & 141.66G & 0.594 & 0.182 & 0.608 & 0.044 & 0.238 & 0.167 & 0.539 & 0.220 & 0.553 & 0.145 & 0.309 & 0.212 \\
PUENet \cite{zhang2023predictive} & 2023-TIP & Image & ViT & 163.16G & 0.594 & 0.204 & 0.619 & 0.037 & 0.302 & 0.212 & 0.673 & 0.427 & 0.803 & 0.034 & 0.499 & 0.389 \\
\noalign{\vspace{2pt}}
\hline
\noalign{\vspace{2pt}}
RCRNet \cite{yan2019semi} & 2019-ICCV & Video & ResNet-50 & 203.14G & 0.597 & 0.174 & 0.583 & 0.025 & 0.194 & 0.137 & - & - & - & - & - & - \\
PNS-Net \cite{ji2021progressively} & 2021-MICCAI & Video & Res2Net-50 & 22.18G & 0.576 & 0.134 & 0.562 & 0.038 & 0.189 & 0.133 & 0.678 & 0.369 & 0.720 & 0.043 & 0.409 & 0.308 \\
MG \cite{yang2021self} & 2021-ICCV & Video & VGG-style & 353.1G & 0.547 & 0.165 & 0.537 & 0.095 & 0.197 & 0.141 & 0.484 & 0.314 & 0.558 & 0.370 & 0.351 & 0.260 \\
SLT-Net \cite{cheng2022implicit} & 2022-CVPR & Video & PVT & 58.18G & 0.656 & 0.357 & 0.785 & 0.021 & 0.387 & 0.310 & 0.679 & 0.420 & 0.805 & 0.033 & 0.445 & 0.342 \\
TSP-SAM \cite{hui2024endow} & 2024-CVPR & Video & SAM & 1414.12G & 0.673 & 0.400 & 0.766 & 0.012 & 0.421 & 0.345 & 0.681 & 0.500 & \textbf{0.853} & 0.031 & 0.496 & 0.393 \\
IMEX \cite{hui2024implicit} & 2024-TMM & Video & ResNet-50 & - & 0.661 & 0.371 & 0.778 & 0.020 & 0.409 & 0.319 & 0.684 & 0.452 & 0.813 & 0.033 & 0.469 & 0.370 \\
EMIP \cite{zhang2025explicit} & 2025-TIP & Video & PVT & 79.29G & 0.675 & 0.381 & - & 0.015 & 0.426 & 0.333 & 0.719 & 0.514 & - & \textbf{0.028} & 0.536 & 0.425 \\
Vcamba (ours) & - & Video & VMamba & \textbf{16.49G} & \textbf{0.684} & \textbf{0.382} & \textbf{0.804} & \textbf{0.010} & \textbf{0.459} & \textbf{0.369} & \textbf{0.729} & \textbf{0.573} & 0.842 & 0.034 & \textbf{0.634} & \textbf{0.509} \\
\noalign{\vspace{2pt}}
\hline
\end{tabular}}
\label{tab1}
\end{table*}

\begin{figure*}[t]
\centerline{\includegraphics[scale=0.50, width=1\linewidth]{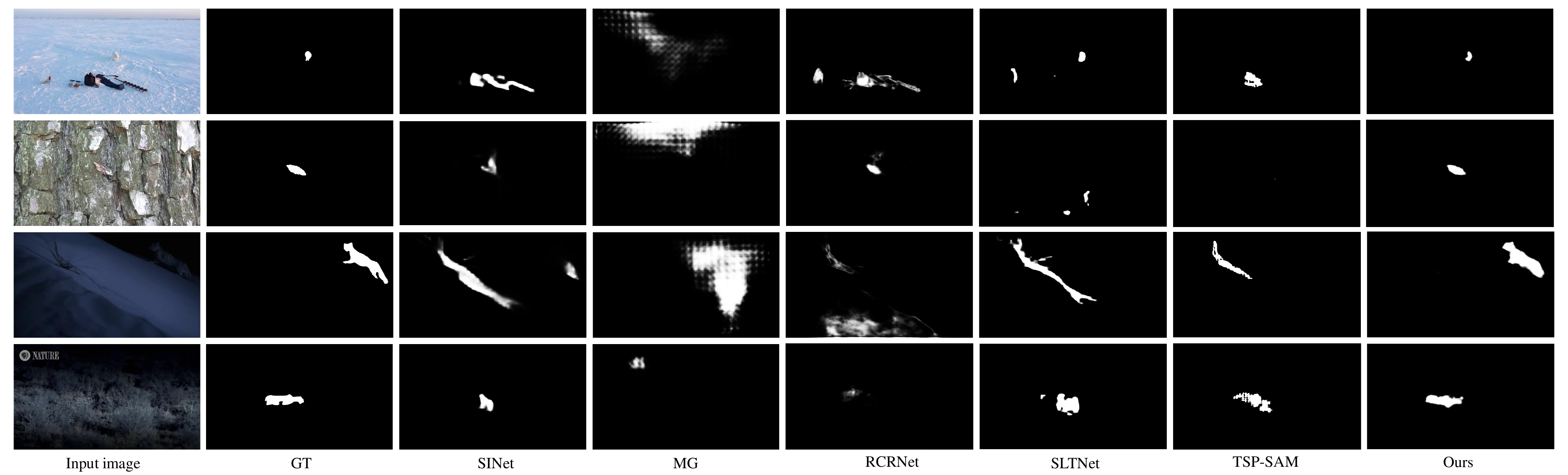}}
\caption{Visual comparison of our method with other five state-of-the-art COD
methods. From left to right: input frame (1st column), GT (2nd column), predicted masks of other methods (3rd-7th columns), and our predicted masks of our Vcamba.}
\label{fig8}
\end{figure*}

\section{EXPERIMENTS AND RESULTS}
\subsection{Datasets and Experiment Settings}
\textbf{Datasets.} Following existing VCOD methods \cite{cheng2022implicit,hui2024implicit,hui2024endow,zhang2025explicit}, the backbone is first pretrained on the COD image dataset COD10K \cite{fan2021concealed}. Subsequently, the entire model is trained on the MoCA-Mask \cite{cheng2022implicit} training set, and tested on both the MoCA-Mask test set and the CAD \cite{bideau2016s} dataset. COD10K is widely used in COD studies \cite{guan2024sdrnet,chen2022boundary,li2025camouflaged}, containing 3,040 training images which is currently the largest publicly available dataset for COD. The MoCA-MASK dataset includes 87 video sequences, with 71 sequences (19,313 frames) for training and 16 sequences (3,626 frames) for testing. Every fifth frame is manually annotated with bounding box and pixel-level segmentation mask. The camouflaged animal dataset (CAD) is a small dataset of camouflaged animals, comprising 9 video sequences and a total of 836 frames. Pixel-level segmentation masks are manually annotated on every fifth frame. In our experiments, the entire CAD dataset is used as a test set.

\textbf{Metric.} Following existing VCOD methods \cite{cheng2022implicit,hui2024implicit,hui2024endow,zhang2025explicit}, we evaluate Vcamba on six standard evaluation metrics: S-measure($S_{\alpha}$) \cite{fan2017structure}, weighted F-measure($F_{\beta}^{\omega}$) \cite{margolin2014evaluate}, mean Enhanced-alignment measure($E_{\phi}$) \cite{fan2021cognitive}, mean absolute error($M$) \cite{perazzi2012saliency}, meanDice ($mDice$), meanIoU ($mIoU$). The better performance corresponds to higher $S_{\alpha}$, $F_{\beta}^{\omega}$, $E_{\phi}$, $mDice$, $mIoU$, and lower $M$.

\textbf{Implementation Details.} We implement our model using PyTorch on a NVIDIA GeForce RTX 4090 (24G) GPU. All images are resized to $386\times 386$. For both training and pretraining data, we apply data augmentation strategies including random cropping, random flipping, and color enhancement. The model adopts VMamba \cite{liu2024vmamba} as the backbone, which is first pretrained on the training set of COD10K. Pretraining takes 8 hours for 100 epochs with a batch size of 6, and backbone parameters are optimized using Adam with a learning rate of 1e-5, reduced by a factor of 10 every 50 epochs. The model is then trained on the MoCA-MASK training set, initializing with the pretrained weights. Training takes 2.5 hours for 10 epochs, using a batch size of 2, and the Adam optimizer with a learning rate of 1e-6.

\subsection{Performance Comparisons}
\textbf{Baselines.} We evaluate our model on two datasets against 14 state-of-the-art methods. These methods can be categorized into two categories: (i) The image-based COD methods including SINet \cite{fan2020camouflaged}, SINet-v2 \cite{fan2021concealed}, ZoomNet \cite{pang2022zoom}, BGNet \cite{sun2022bgnet}, FEDERNet \cite{he2023camouflaged}, FSPNet \cite{huang2023feature}, and PUENet \cite{zhang2023predictive}. (ii) Video object segmentation methods including RCRNet \cite{yan2019semi}, PNS-Net \cite{ji2021progressively}, MG \cite{yang2021self}, SLT-Net \cite{cheng2022implicit}, TSP-SAM \cite{hui2024endow}, IMEX \cite{hui2024implicit}, and EMIP \cite{zhang2025explicit}. All the methods are representative works that are widely used as baselines in VCOD. To ensure fair evaluation, their results are either obtained from official releases or generated using publicly available source codes.

\textbf{Quantitative Results.}
Table~\ref{tab1} presents quantitative comparisons between our method and existing methods on MoCA-MASK and CAD datasets. The results show that our method outperforms them. Specifically, compared to representative and novel static image-based methods such as SINet and PUENet, our approach achieves significant improvements across all metrics on both datasets, confirming the value of motion cues for camouflage breaking and the effectiveness of Mamba in implicitly modeling multi-frame motion. 

Compared to the latest and representative VCOD methods, such as SLTNet and EMIP, our method demonstrates substantial superiority across most metrics. Specifically, on MoCA-MASK, it achieves improvements of 4.3\% ($S_{\alpha}$), 7.0\% ($F_{\beta}^{\omega}$), 2.4\% ($E_{\phi}$), 52.3\% ($M$), 18.6\% ($mDice$) and 19\% ($mIoU$) over SLTNet. On CAD, the performance gains reach 7.3\% ($S_{\alpha}$), 36.4\% ($F_{\beta}^{\omega}$), 4.6\% ($E_{\phi}$), 42.5\% ($mDice$) and 48.8\% ($mIoU$). Furthermore, compared to the latest method EMIP, our approach shows improvements of 1.3\% and 1.4\% in $S_{\alpha}$ on both datasets, demonstrating the dual-domain perception and motion fusion achieve more complete camouflaged object localization. The significant advantages in $F_{\beta}^{\omega}$, $mDice$ and $mIoU$ further validate the effectiveness of multi-scale receptive fields and adaptive frequency enhancement in improving local detail segmentation.

Notably, while SLTNet employs transformers for image encoding and long-range motion perception, IMEX utilizes ResNet50 \cite{he2016deep}, and TSP-SAM leverages the Segment Anything Model (SAM) with powerful generalization capability, our approach outperforms all these methods, which not only validates Mamba's feasibility for VCOD but also demonstrates its superiority over transformers and CNNs in sequence modeling and motion perception.  Furthermore, as shown in Table~\ref{tab1}, we further report the MACs of our model, and compare them with representative open-source baselines. These results confirm that our Mamba-based method achieves not only stronger performance but also higher computational efficiency due to its linear time complexity.

\textbf{Qualitative Results.}
As shown in Fig.~\ref{fig8}, we present the predictions of our method alongside representative baselines such as SLTNet, PUENet, and FEDERNet across a set of typical scenes. Results demonstrate our predictions exhibit superior global consistency with ground truth, achieving more complete camouflaged object localization, and more accurate local details, particularly in object boundaries. These observations validate the importance of constructing dual-domain motion perception and fusion design from both global and local perspectives. Notably, our method maintains robust detection in complex scenes where other methods show false positives or missed detections, further confirming its robustness.

\begin{table}[t]
\caption{ABLATION STUDY OF CORE COMPONENTS ON MOCA-MASK DATASET}
\centering
\fontsize{6}{7}\selectfont
\resizebox{\linewidth}{!}{
\begin{tabular}{c|ccccc} 
\hline
Method & {\cellcolor[rgb]{0.89,0.89,0.89}}$S_{\alpha}\uparrow$ & {\cellcolor[rgb]{0.89,0.89,0.89}}$F_{\beta}^{\omega}\uparrow$ & {\cellcolor[rgb]{0.89,0.89,0.89}}$E_{\phi}\uparrow$ & 
{\cellcolor[rgb]{0.89,0.89,0.89}}$mDice\uparrow$ & {\cellcolor[rgb]{0.89,0.89,0.89}}$mIoU\uparrow$ \\ 
\hline
\noalign{\vspace{2pt}}
A1 & 0.679 & 0.372 & 0.812 & 0.445 & 0.358 \\ 
A2 & 0.681 & 0.374 & 0.799 & 0.451 & 0.363 \\ 
A3 & 0.677 & 0.367 & 0.792 & 0.444 & 0.356 \\ 
A4 & 0.677 & 0.370 & 0.800 & 0.444 & 0.357 \\ 
A5 & 0.676 & 0.372 & 0.797 & 0.445 & 0.357 \\ 
A6 & 0.678 & 0.369 & 0.791 & 0.447 & 0.358 \\ 
A7 & 0.675 & 0.365 & 0.765 & 0.444 & 0.356 \\ 
Vcamba & 0.684 & 0.382 & 0.804 & 0.459 & 0.370 \\ 
\noalign{\vspace{2pt}}
\hline
\end{tabular}}
\label{tab3}
\end{table}

\subsection{Ablation Studies}
To thoroughly evaluate the contributions of the core components and demonstrate their superiority over traditional modules, we conducted comprehensive ablation studies on the larger MoCA-MASK test set, which provides a more robust benchmark than CAD.

\textbf{Effectiveness of network modules.} As demonstrated in Table~\ref{tab3}, systematic module-wise ablation experiments are performed by removing individual components from Vcamba (denoted as: A1: without RFVSS, A2: without AFE, A3: without DSE, A4: without SLMP, A5: without FLMP, A6: without SFMF). The results demonstrate that removing any single module leads to a noticeable performance drop, clearly validating the effectiveness of each component in VCOD.

Specifically, to assess the complementary role of frequency in improving discriminability of spatial appearance, we conduct a comparative experiment by completely removing the frequency branch, retaining only the spatial branch (denoted as A7 in Table~\ref{tab3}). In this setting, the model solely relies on spatial-domain features to extract camouflaged cues and perceive motion information. The results reveal a sharp performance decline, with key metrics decreasing by 1.3\%($S_{\alpha}$), 4.6\%($F_{\beta}^{\omega}$) and 5.1\%($E_{\phi}$). This empirically demonstrates that frequency features not only provide complementary discriminative cues but also offer global context and motion information that help preserve object structural integrity. These results also demonstrate that combining spatial and frequency domains enables more comprehensive feature representations.

\begin{table}[t]
\caption{Ablation study of different backbones.}
\centering
\fontsize{6}{7}\selectfont
\resizebox{0.8\linewidth}{!}{
\begin{tabular}{c|c|ccc} 
\hline
Method & {\cellcolor[rgb]{0.89,0.89,0.89}}$MACs$ & {\cellcolor[rgb]{0.89,0.89,0.89}}$S_{\alpha}\uparrow$ & {\cellcolor[rgb]{0.89,0.89,0.89}}$F_{\beta}^{\omega}\uparrow$ & {\cellcolor[rgb]{0.89,0.89,0.89}}$mIoU\uparrow$ \\ 
\hline
\noalign{\vspace{2pt}}
PVTv2-B5 & 91.24G & 0.676 & 0.362 & 0.364 \\ 
ViT-B & 139.59G & 0.641 & 0.304 & 0.294 \\ 
Res2Net-50 & 45.75G & 0.584 & 0.199 & 0.200 \\
VMamba (Ours) & 16.49G & 0.684 & 0.382 & 0.369 \\ 
\noalign{\vspace{2pt}}
\hline
\end{tabular}}
\label{tab7}
\end{table}

\textbf{Effectiveness of VMamba.} VMamba enables global sequence modeling with linear time complexity, while its selective scanning allows the model to adaptively focus on critical regions. Compared with other visual Mambas, VMamba explicitly adopts a cross-scan mechanism to model long-range dependencies across both horizontal and vertical directions, making it better suited for 2D visual representations. To validate the effectiveness of it, as shown in Table~\ref{tab7}, we replace the backbone with PVTv2-B5 \cite{wang2022pvt}, Res2Net \cite{gao2019res2net} and ViT-B \cite{dosovitskiy2020image}, and conduct experiments under identical training strategies. VMamba achieves the best performance across all metrics with lower computational cost, indicating its superior performance–efficiency trade-offs in VCOD.

\begin{table}[t]
\caption{ABLATION STUDY OF FREQUENCY LEARNING METHODS IN AFE}
\centering
\fontsize{6}{7}\selectfont
\resizebox{\linewidth}{!}{
\begin{tabular}{c|ccccc} 
\hline
Method & {\cellcolor[rgb]{0.89,0.89,0.89}}$S_{\alpha}\uparrow$ & {\cellcolor[rgb]{0.89,0.89,0.89}}$F_{\beta}^{\omega}\uparrow$ & {\cellcolor[rgb]{0.89,0.89,0.89}}$E_{\phi}\uparrow$ & 
{\cellcolor[rgb]{0.89,0.89,0.89}}$mDice\uparrow$ & {\cellcolor[rgb]{0.89,0.89,0.89}}$mIoU\uparrow$ \\ 
\hline
\noalign{\vspace{2pt}}
Conv$3\times 3$ & 0.676 & 0.374 & 0.806 & 0.447 & 0.359 \\ 
Self-Attention & 0.680 & 0.375 & 0.809 & 0.441 & 0.356 \\ 
Cross Scan & 0.679 & 0.376 & 0.787 & 0.455 & 0.365 \\ 
Ours & 0.684 & 0.382 & 0.804 & 0.459 & 0.370 \\ 
\noalign{\vspace{2pt}}
\hline
\end{tabular}}
\label{tab4}
\end{table}

\textbf{Effectiveness of AFE.} Our Mamba-based AFE module innovatively achieves adaptive frequency component enhancement through a frequency-domain sequential scanning (FSS) strategy. To further validate the effectiveness of this design, we replace the Mamba block in AFE with conventional frequency learning methods such as convolution $3\times 3$ and self-attention, and we also replace our FSS with the cross scan of VMamba, as it is the most representative and performant scanning strategy among existing Mamba variants. The results shown in Table~\ref{tab4} demonstrate the advantages of our method in frequency learning. Furthermore, the proposed FSS outperforms mainstream cross scan in frequency map serialization, as it effectively preserves the continuity, causality, and progressive structural evolution inherent in frequency component superposition after unfolding them into ordered sequences.

\begin{table}[t]
\caption{ABLATION STUDY OF LONG-RANGE MOTION PERCEPTION METHODS IN SLMP AND FLMP.}
\centering
\fontsize{6}{7}\selectfont
\resizebox{\linewidth}{!}{
\begin{tabular}{c|ccccc} 
\hline
Method & {\cellcolor[rgb]{0.89,0.89,0.89}}$S_{\alpha}\uparrow$ & {\cellcolor[rgb]{0.89,0.89,0.89}}$F_{\beta}^{\omega}\uparrow$ & {\cellcolor[rgb]{0.89,0.89,0.89}}$E_{\phi}\uparrow$ & 
{\cellcolor[rgb]{0.89,0.89,0.89}}$mDice\uparrow$ & {\cellcolor[rgb]{0.89,0.89,0.89}}$mIoU\uparrow$ \\ 
\hline
\noalign{\vspace{2pt}}
3D CNN & 0.678 & 0.380 & 0.818 & 0.451 & 0.362 \\ 
ConvLSTM & 0.674 & 0.362 & 0.774 & 0.434 & 0.344 \\ 
ConvGRU & 0.677 & 0.376 & 0.812 & 0.448 & 0.358 \\ 
LSTM & 0.535 & 0.113 & 0.699 & 0.191 & 0.121 \\ 
Ours & 0.684 & 0.382 & 0.804 & 0.459 & 0.370 \\ 
\noalign{\vspace{2pt}}
\hline
\end{tabular}}
\label{tab5}
\end{table}

\begin{figure}[t]
\centerline{\includegraphics[scale=1, width=1\linewidth]{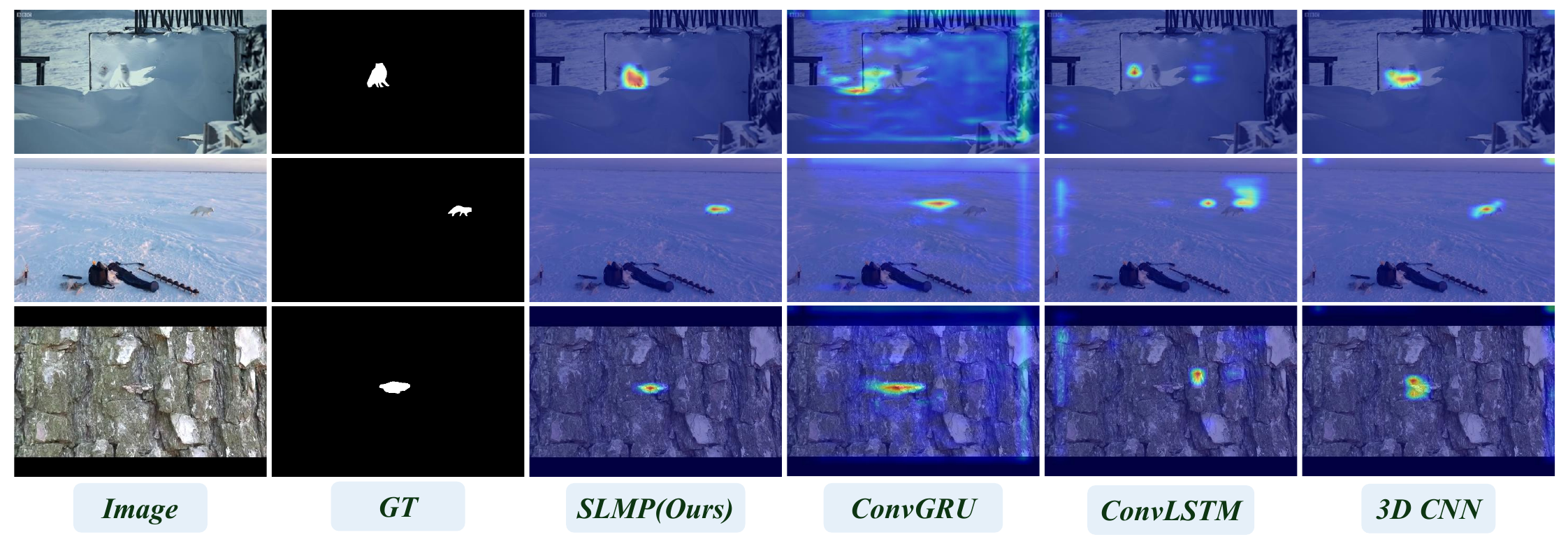}}
\caption{Intermediate feature heatmaps of different long-range motion perception methods in the spatial-domain branch.}
\label{fig9}
\end{figure}

\textbf{Effectiveness of SLMP and FLMP.} This work presents an innovative Mamba-based dual-domain motion perception approach that simultaneously captures spatial and frequency-domain motion cues. As shown in Table~\ref{tab5}, we replace SLMP and FLMP with traditional long-range motion modeling methods such as 3D CNN, ConvLSTM \cite{shi2015convolutional}, and ConvGRU \cite{2015Delving}. Experimental results demonstrate that our Mamba-based modules outperform them. Furthermore, as shown in Fig.~\ref{fig9}, we visualize the intermediate motion features generated by different long-sequence motion modeling methods through heatmaps. The visualization demonstrates that the motion features modeled by SLMP exhibit a more precise target localization and less background noise interference. Both quantitative and qualitative experimental results clearly demonstrate the advantages of our SSM-based motion perception through spatio-temporal and frequency-temporal scanning strategies.

\begin{table}[t]
\caption{Ablation study of different motion modeling strategies in FLMP.}
\centering
\fontsize{6}{7}\selectfont
\resizebox{\linewidth}{!}{
\begin{tabular}{c|cc|ccc} 
\hline
Method & {\cellcolor[rgb]{0.89,0.89,0.89}} Carrier & {\cellcolor[rgb]{0.89,0.89,0.89}} Guidance & {\cellcolor[rgb]{0.89,0.89,0.89}}$S_{\alpha}\uparrow$ & {\cellcolor[rgb]{0.89,0.89,0.89}}$F_{\beta}^{\omega}\uparrow$ & {\cellcolor[rgb]{0.89,0.89,0.89}}$mIoU\uparrow$ \\ 
\hline
\noalign{\vspace{2pt}}
Amplitude-only & Amplitude Spectrum & \ding{52} & 0.679 & 0.377 & 0.362 \\ 
Amplitude + Phase & Frequency Spectrum & \ding{55} & 0.679 & 0.375 & 0.358 \\ 
No Guidance & Phase Spectrum & \ding{55} & 0.680 & 0.376 & 0.357 \\
Phase-only (Ours) & Phase Spectrum & \ding{52} & 0.684 & 0.382 & 0.369 \\
\noalign{\vspace{2pt}}
\hline
\end{tabular}}
\label{tab8}
\end{table}

\textbf{Effectiveness of Phase-Based Motion Modeling.} In the FLMP, we explicitly separate the spectral phase and magnitude. Motion is perceived by modeling temporal variations of phase, which is further used to guide amplitude filtering. As shown in Table~\ref{tab8}, we conduct ablation studies comparing phase-only motion modeling with amplitude-only and phase–amplitude combination (i.e., frequency spectrum) strategies. The results indicate that the phase-only strategy achieves the best performance, whereas the other variants suffer performance degradation due to the introduction of additional noise. This observation aligns with the Fourier shift theorem, where phase variations directly correspond to spatial displacement and are more robust to appearance changes. Moreover, directly combining original amplitude with phase leads to performance drops, confirming the necessity of phase-guided amplitude filtering for noise suppression.

\begin{table}[t]
\caption{ABLATION STUDY OF MOTION FEATURE FUSION METHODS IN SFMF.}
\centering
\fontsize{6}{7}\selectfont
\resizebox{\linewidth}{!}{
\begin{tabular}{c|ccccc} 
\hline
Method & {\cellcolor[rgb]{0.89,0.89,0.89}}$S_{\alpha}\uparrow$ & {\cellcolor[rgb]{0.89,0.89,0.89}}$F_{\beta}^{\omega}\uparrow$ & {\cellcolor[rgb]{0.89,0.89,0.89}}$E_{\phi}\uparrow$ & 
{\cellcolor[rgb]{0.89,0.89,0.89}}$mDice\uparrow$ & {\cellcolor[rgb]{0.89,0.89,0.89}}$mIoU\uparrow$ \\ 
\hline
\noalign{\vspace{2pt}}
Conv$3\times 3$ & 0.679 & 0.377 & 0.810 & 0.447 & 0.361 \\ 
Cross-Attention & 0.671 & 0.359 & 0.776 & 0.441 & 0.349 \\ 
Ours & 0.684 & 0.382 & 0.804 & 0.459 & 0.370 \\ 
\noalign{\vspace{2pt}}
\hline
\end{tabular}}
\label{tab6}
\end{table}

\begin{figure}[b]
\centerline{\includegraphics[scale=1, width=1\linewidth]{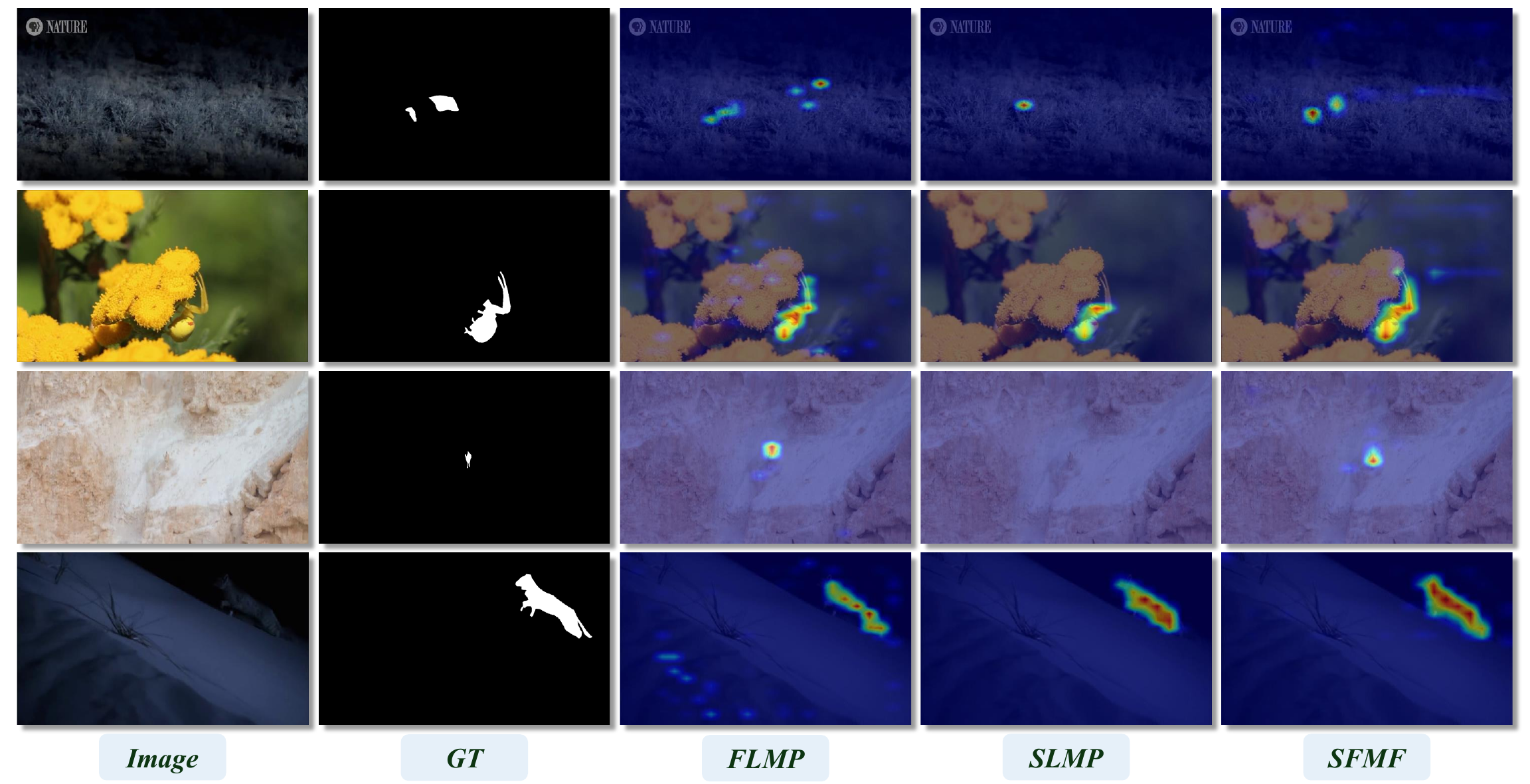}}
\caption{Visual comparisons of spatial motion feature through SLMP, frequency motion feature through FLMP and unified motion representation through SFMF.}
\label{fig10}
\end{figure}

\textbf{Effectiveness of SFMF.} We introduce SFMF, a novel Mamba-based dual-domain motion feature fusion module that employs a dual-domain sequence concatenation strategy in sequence-to-sequence and point-to-point manners for both global cross-domain and local structural integration. Fig.~\ref{fig10} presents heatmaps of the motion features processed by FLMP (frequency domain) and SLMP (spatial domain), along with their fused representation through SFMF. The results demonstrate SFMF's effectiveness in complementary feature integration and domain-specific noise suppression. As we can see, in the first row, while FLMP locates only one target with significant background noise and SLMP detects the other target, their fusion successfully identifies both targets while eliminating domain-specific noise. The third row particularly highlights the critical advantage of frequency, when spatial features completely fail to detect the target, frequency-based motion perception through energy dynamics maintains reliable detection, which strongly justifies our dual-domain design.

\begin{figure}[t]
\centerline{\includegraphics[scale=1, width=1\linewidth]{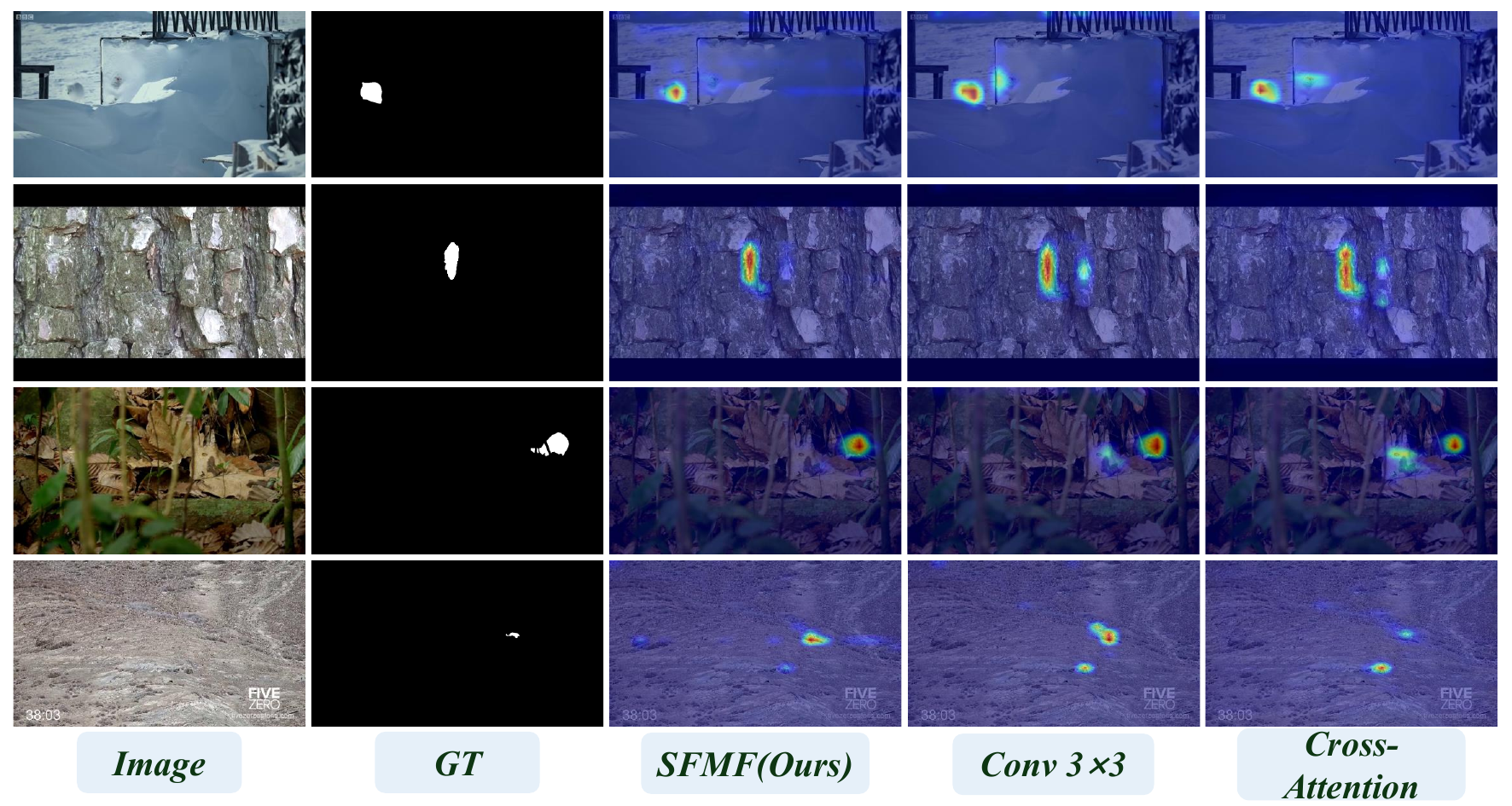}}
\caption{Visual comparisons of unified motion representations through different motion feature fusion methods.}
\label{fig11}
\end{figure}

To further validate the superiority of SFMF, as shown in Table~\ref{tab6}, we replace SFMF with convolution $3\times 3$ and cross-attention, the results show that SFMF achieves better performance than traditional methods. Furthermore, Fig.~\ref{fig11} demonstrates that our SFMF module effectively captures local motion details, and globally integrating cross-domain context to filter irrelevant noise through its point-to-point and sequence-to-sequence fusion strategy.

\subsection{Limitation}
While Mamba theoretically reduces computational complexity via linear-time sequence modeling, this advantage does not totally translate into superior runtime performance in practical deployments. Compared to highly optimized self-attention on GPUs, Mamba’s selective scan relies on hardware- and implementation-specific optimizations \cite{gu2023mamba,pmlrv235dao24a,asif2025perfmamba,liu2024vmamba}. Existing general visual Mamba, such as VMamba \cite{liu2024vmamba} and Vision Mamba \cite{pmlrv235zhu24f}, do not exhibit a clear advantage in inference speed (FPS) over well-optimized Transformers. In our experiments, Vcamba achieves 14 FPS with VMamba, comparable to 13 FPS with PVT, suggesting that GPU optimization significantly constrains the practical efficiency of visual Mamba. Further hardware-aware optimizations are therefore necessary to fully exploit its theoretical advantages.

\section{CONCLUSION}
In this work, we propose a dual-domain motion perception network called Vcamba for VCOD that leverages both spatial and frequency cues. Frequency features are introduced to enhance spatial discriminability and global perception. We find that frequency components exhibit a structural evolution pattern during ordered superposition, forming a causal sequence that can be effectively modeled by SSMs. Based on this insight, we introduce an adaptive frequency enhancement (AFE) module with a novel frequency-domain sequential scanning (FSS) strategy to enable effective frequency learning. To further exploit motion cues, we design space-based and frequency-based long-range motion perception (SLMP, FLMP) modules to model spatio-temporal and frequency-temporal dependencies for dual-domain motion perception. A spatial and frequency motion fusion (SFMF) module is finally employed to unify complementary motion cues via our dual-domain sequence concatenation strategy. Extensive experiments show that Vcamba outperforms state-of-the-art methods on two datasets with lower computational cost, confirming its effectiveness and efficiency.

\bibliographystyle{IEEEtran}
\bibliography{refs}
\end{document}